\documentclass{article}

\usepackage[nonatbib,preprint]{neurips_data_2022}

\usepackage[utf8]{inputenc} 
\usepackage[T1]{fontenc}    
\usepackage[hidelinks]{hyperref}       
\usepackage{url}            
\usepackage{booktabs}       
\usepackage{amsfonts}       
\usepackage{nicefrac}       
\usepackage{microtype}      
\usepackage{xcolor}         
\usepackage{wrapfig}        
\usepackage{amssymb,amsmath}
\usepackage{subcaption}
\usepackage{listings}
\usepackage{tikz}

\usepackage{pgfplots}
\pgfplotsset{
  compat = 1.16, 
}

\definecolor{bleu}     {RGB}{117,112,179}
\definecolor{cardinal} {RGB}{217,95,2}
\definecolor{emerald}  {RGB}{27,158,119}
\definecolor{lightgrey}{RGB}{230,230,230}
\definecolor{burgundy}{RGB}{	185, 14, 10}

\usepackage[%
commentmarkup = uwave,
defaultcolor=burgundy
]{changes}

\hypersetup{pdfinfo={
		Title={All the World's a (Hyper)Graph: A Data Drama},
		Author={Corinna Coupette, Jilles Vreeken, and Bastian Rieck}
}}

\usepackage[
	backend=biber,
	sortcites=true
]{biblatex}
\bibliography{main.bib}

\title{All the World's a (Hyper)Graph:\\A Data Drama}

\author{%
  Corinna Coupette\\
  Max Planck Institute\\ for Informatics\\
  Saarbrücken, Germany\\
  \And
  Jilles Vreeken\\
  CISPA Helmholtz Center\\ for Information Security\\
  Saarbrücken, Germany\\
   \And
   Bastian Rieck \\
   Institute of AI for Health\\
   Helmholtz Munich \\
   Munich, Germany\\
}

\usepackage{pdflscape}
\usepackage{multirow}
\usepackage{xcolor,xspace}
\usepackage{enumitem}

\newcommand{\ourdata}{\textsc{Hyperbard}\xspace}

\newcounter{playlinenumber}
\newcommand{\pll}[1]{\stepcounter{playlinenumber}\scriptsize{\theplaylinenumber}&#1\\}
\newcommand{\prl}[1]{\stepcounter{playlinenumber} #1&\scriptsize{\theplaylinenumber}\\}
\newcommand{\pln}[1]{&#1\\}
\newcommand{\prn}[1]{#1&\\}
\newcommand{\pls}[1]{&\emph{#1}\\}
\newcommand{\prs}[1]{\emph{#1}&\\}

\newcommand{\playauthors}{\textsc{Authors}\xspace}
\newcommand{\playauthorsabbr}{\emph{Auth.}\xspace}
\newcommand{\reviewer}{\textsc{Reviewer}\xspace}
\newcommand{\reviewerabbr}{\emph{Rev.}\xspace}
\newcommand{\professor}{\textsc{Professor}\xspace}
\newcommand{\professorabbr}{\emph{Prof.}\xspace}
\newcommand{\secretary}{\textsc{Secretary}\xspace}
\newcommand{\secretaryabbr}{\emph{Sec.}\xspace}
\newcommand{\seniorresearcher}{\textsc{Senior Researcher}\xspace}
\newcommand{\seniorresearcherabbr}{\emph{Sen.\thinspace R.}\xspace}
\newcommand{\tutor}{\textsc{Tutor}\xspace}
\newcommand{\tutorabbr}{\emph{Tut.}\xspace}
\newcommand{\creature}{\textsc{Creature}\xspace}
\newcommand{\creatureabbr}{\emph{Cre.}\xspace}
\newcommand{\colleague}{\textsc{Colleague}\xspace}
\newcommand{\colleagueabbr}{\emph{Col.}\xspace}
\newcommand{\deadlines}{\textsc{Deadlines}\xspace}

\newcommand{\deadlineone}{\textsc{First Deadline}\xspace}
\newcommand{\deadlineoneabbr}{\emph{First Dea.}\xspace}
\newcommand{\deadlinetwo}{\textsc{Second Deadline}\xspace}
\newcommand{\deadlinetwoabbr}{\emph{Sec. Dea.}\xspace}

\newcommand{\deadlinethreeabbr}{\emph{Third Dea.}\xspace}
\newcommand{\hyperbard}{\textsc{Hyperbard}\xspace}
\newcommand{\hyperbardabbr}{\emph{Hyp.}\xspace}
\newcommand{\graph}{\textsc{Graph}\xspace}
\newcommand{\graphabbr}{\emph{Gra.}\xspace}
\newcommand{\randj}{\emph{Romeo and Juliet}\xspace}

\newcommand{\datarepo}{Zenodo\xspace}
\newcommand{\datapage}{\href{https://hyperbard.net}{https://hyperbard.net}\xspace}
\newcommand{\coderepo}{GitHub\xspace}
\newcommand{\coderepourl}{\href{https://github.com/hyperbard/hyperbard}{https://github.com/hyperbard/hyperbard}\xspace}
\newcommand{\docsurl}{\href{https://hyperbard.readthedocs.io/en/latest/}{https://hyperbard.readthedocs.io/en/latest/}\xspace}
\newcommand{\tutorialrepourl}{\href{https://github.com/hyperbard/tutorials}{https://github.com/hyperbard/tutorials}\xspace}
\newcommand{\binderurl}{\href{https://mybinder.org/v2/gh/hyperbard/sandbox/main?urlpath=git-pull\%3Frepo\%3Dhttps\%253A\%252F\%252Fgithub.com\%252Fhyperbard\%252Ftutorials\%26urlpath\%3Dlab\%252Ftree\%252Ftutorials\%252Fnotebooks\%252Fwelcome.ipynb\%26branch\%3Dmain}{Binder}\xspace}
\newcommand{\ourdatalicense}{\href{https://creativecommons.org/licenses/by-nc/4.0/}{CC BY-NC 4.0}\xspace}
\newcommand{\ourdatadoi}{\href{https://doi.org/10.5281/zenodo.6627159}{10.5281/zenodo.6627159}}
\newcommand{\ourdatadoiall}{\href{https://doi.org/10.5281/zenodo.6627158}{10.5281/zenodo.6627158}}
\newcommand{\ourcodelicense}{\href{https://github.com/hyperbard/hyperbard/blob/main/LICENSE}{BSD 3-Clause}\xspace}
\newcommand{\ourcodedoi}{\href{https://doi.org/10.5281/zenodo.6627161}{10.5281/zenodo.6627161}}
\newcommand{\ourcodedoiall}{\href{https://doi.org/10.5281/zenodo.6627158}{10.5281/zenodo.6627160}}
\newcommand{\folgerlicense}{\href{https://creativecommons.org/licenses/by-nc/3.0/}{CC BY-NC 3.0 Unported}\xspace}
\newcommand{\folgerlibrary}{Folger Shakespeare Library\xspace}
\newcommand{\folgertexts}{Folger Digital Texts\xspace}
\newcommand{\milena}{graph learning, graph mining, and network analysis\xspace}
\newcommand{\cc}{Corinna Coupette\xspace}
\newcommand{\br}{Bastian Rieck\xspace}

\newcommand{\checkednumber}[1]{#1}

\NewBibliographyString{dataset}
\NewBibliographyString{code}

\DefineBibliographyStrings{english}{
	dataset = {dataset},
	code = {code}
}

\renewbibmacro{begentry}{%
	\ifkeyword{meta}{\textsuperscript{*}}{}%
	\ifentrytype{dataset}
	{\bibstring[\mkbibbrackets]{dataset}%
		\setunit{\addspace}}
	{}%
	\ifentrytype{code}
	{\bibstring[\mkbibbrackets]{code}%
		\setunit{\addspace}}
	{}%
}

\graphicspath{{./figures/}}

\begin{document}

\maketitle
\phantomsection
\addcontentsline{toc}{section}{Abstract}
\begin{abstract}

We introduce \ourdata, a dataset of diverse relational data representations derived from Shakespeare's plays. 
Our representations range from simple \emph{graphs} capturing character
co-occurrence in single scenes to \emph{hypergraphs} encoding complex communication settings and character contributions as hyperedges with edge-specific node weights.
By making multiple intuitive representations readily available for experimentation, 
we facilitate rigorous \emph{representation robustness checks} in graph learning, graph mining, and network analysis,
highlighting the advantages and drawbacks of specific representations.
Leveraging the data released in \ourdata, we demonstrate that many solutions to popular graph mining problems are highly dependent on the representation choice, thus calling current graph curation practices into question.
As an homage to our data source, and asserting that science can also be art, we present all our points in the form of a play.\footnote{%
	This manuscript is completely AI-free.
	An abridged version was published in \emph{Digital Scholarship in the Humanities} \cite{coupette2023dsh}.
}
\end{abstract}

\vspace*{2em}

\phantomsection
\addcontentsline{toc}{section}{Dramatis Person\ae}
\begin{figure}[h]\small
	\centering \hspace*{1em}DRAMATIS PERSON\AE\\[6pt]
	\hspace*{-3.15em}\begin{tabular}[t]{rp{0.5\linewidth}|}
		\phantom{\scriptsize{1}}&\playauthors.\multirow{2}{*}{\hspace*{4.5em}$\Big\}$ Persons in the Induction.}\\
		\pln{\reviewer, a reader.}
		\pln{\creature, a curious mind.}
		\pln{\hyperbard, a faun, sovereign of spirits.}
		\pln{\graph, a gentle spirit.}
		\pln{}
	\end{tabular}~
	\begin{tabular}[t]{p{0.5\linewidth}r}
		\professor,\multirow{3}{0.7\linewidth}{\hspace*{4.5em}$\Bigg\}$ Part of the Community.}&\\
		\prn{\seniorresearcher,}
		\prn{\colleague.}
		\tutor,\multirow{3}{0.8\linewidth}{\hspace*{4.5em}$\Bigg\}$ Serving the Community.}&\\
		\prn{\secretary,}
		\prn{\deadlines.}
	\end{tabular}\vspace*{15pt}
	
	\textsc{Scene.}---\emph{Sometimes in the Community; and sometimes in the forest.}\vspace*{7pt}
	
	\rule{10em}{0.4pt}
\end{figure}

\begin{figure}[h]\small
	\hspace*{-3.15em}\begin{tabular}[t]{rp{0.5\linewidth}|}
		\pls{\phantomsection\addcontentsline{toc}{section}{Induction}\hfil\textsc{INDUCTION.}}
		\pls{\phantomsection\addcontentsline{toc}{subsection}{Scene I}\hfil\textsc{Scene I.}---Between submission and decision.}
		\pls{Enter \reviewer and \playauthors.}
		\pll{\quad\reviewerabbr What is this? Is this not against the rules?}
		\pll{\quad\playauthorsabbr The columns? These are only simple tables.}
		\pll{They serve to help us implement blank verse.}
		\pll{The script-sized numbers count the spoken lines,}
		\pll{They disappear when folks use prose at times.}
		\pll{We introduce a novel dataset,}
	\end{tabular}
	\begin{tabular}[t]{p{0.5\linewidth}r}
		\prl{With full documentation as Appendix.}
		\prl{Raw data stem from all of Shakespeare's plays \cite{folger2022},}
		\prl{We model them as graphs in many ways,}
		\prl{And demonstrate representations matter.}
		\prl{The data readily accessible \cite{coupette2022hyperdata},}
		\prl{All code is publicly available \cite{coupette2022hypercode}.}
		\prl{What follows, to avoid redundancy,}
		\prl{Conveys our main ideas, as you will see}
		\prl{A tragedy in the Community.}
	\end{tabular}
\end{figure}
\clearpage
\begin{figure}[h]\small
	\hspace*{-3.4em}\begin{tabular}[t]{rp{0.5\linewidth}|}
		\pls{\phantomsection\addcontentsline{toc}{section}{Act I}\hfil \textsc{ACT I.}}
		\pls{\phantomsection\addcontentsline{toc}{subsection}{Scene I}\hfil\textsc{Scene I.}---The Community. \professor's office.}
		\pls{Enter \seniorresearcher and \tutor, bearing a barrow. 
			On the barrow, a swooning \creature, feeble but breathing.}
		\pll{\quad\tutorabbr They must have hit a rock while on our problem.}
		\pll{\quad\seniorresearcherabbr Did they get hurt? Who are they, anyway?}
		\pls{They put down the barrow.} 
		\pls{Enter \professor and \secretary.}
		\pll{\quad \professorabbr What is this fuss? Did they get an appointment?}
		\pll{Another rescue? Do they know to code?}
		\pll{\quad \tutorabbr Should we employ them?}
		\pln{\quad \professorabbr \hspace{7em}What, you mean by contract?}
		\pll{\quad\seniorresearcherabbr It seems that they are really good with graphs.}
		\pll{\quad\professorabbr All right---}
		\pls{\creature moves.}
		\pln{\quad \tutorabbr \hspace{4em}Be quick, they wake!}
		\pln{\quad \secretaryabbr \hfill I'll get the forms.}
		\pls{Exit \secretary.}
		\pls{\creature shuffles, sighs, and sits up.}
		\pll{\quad \professorabbr Welcome to the Community!}
		\pls{\hfill [They smile generously.]}
		\pll{\quad \creatureabbr \hspace{12em}The what?}
		\pll{Where am I---Why is everything so clean? 
			Wasn't I chasing bugs, out in the woods? 
			Or roaming pastures, playing in the mud? 
			I fail to recollect; I must be dreaming. 
			So is this but a nightmare? Or a prison? \addtocounter{playlinenumber}{3}} 
		\pll{Am I a hostage?}
		\pll{\quad \professorabbr \hspace*{3.5em}Fellow, you are free!}
		\pls{Re-enter \secretary, handing \professor the forms.}
		\pll{\quad \professorabbr Just sign here, will you?}
		\pls{They point to a field in the forms. \creature signs.}
		\pln{~~\professorabbr \emph{and} \seniorresearcherabbr \emph{[in synchrony]} Welcome to your PhD!}
		\pls{Exeunt.}
		\pls{}
		\pls{\phantomsection\addcontentsline{toc}{subsection}{Scene II}\hfil\textsc{Scene II.}---\creature's office.}
		\pls{Enter \creature, closing the door. 
			They pace about the room, then settle before the window.\hypertarget{luther}{}}
		\pll{\quad\creatureabbr So here I stand; and I can do no other? 
			Little do I remember of my roots. 
			Well, elsewhere sure they lie, but must I cut them? 
			How will I learn this play, and play my part?}\addtocounter{playlinenumber}{3}
		\pls{Knocking.}
		\pll{\quad\creatureabbr Come in!---\emph{[Aside]} Stay out!}
		\pls{Enter \colleague.}
		\pll{\quad\colleagueabbr Hello, how are you?
			You must be the new one!}
		\pll{You work on graphs, or that's what I've been told?}
		\pll{They said you came from outside, from the forest.}
		\pll{Well, better not go back---here, we do trees.}
		\pll{\quad\creatureabbr What does that mean?}
		\pll{\quad\colleagueabbr We like to operate with clear-cut questions,}
		\pll{Employing very powerful abstractions.}
		\pll{To be successful, publish many units,}
		\pll{At top-ranked venues, making single points.}
		\pll{Evaluate on standard datasets,}
		\pll{And over-promise, then, to hedge your bets.}
		\pll{\quad\creatureabbr So, this is science?}
		\pln{\quad\colleagueabbr \hspace*{8em}It is how things work.}
		\pls{Exit \colleague.}
		\pls{Enter \professor with \deadlines.}
		\pll{\quad\professorabbr So let me introduce you to your guardians.}
		\pll{We call them \deadlines---never mind the name.}
		\pll{They form the circle of scientific life,}
		\pll{And soon will be your greatest motivators.}
		\pll{As papers pave your path to graduation,}
		\pll{Your thinking becomes music to their beat.}
	\end{tabular}~
	\begin{tabular}[t]{p{0.5\linewidth}r}
		\prl{\quad\creatureabbr But why?}
		\prn{\quad\deadlineoneabbr \hspace*{1.5em}Why what?}
		\prn{\quad\deadlinetwoabbr \hspace*{4.5em}It's pressure making diamonds.}
		\prl{\quad\deadlinethreeabbr We set incentives, we're just here to help!}
		\prl{\mbox{\quad\professorabbr I'll leave you with them, then, you'll get familiar.}}
		\prl{And when you're done, make sure to put my name.}
		\prs{Exit. \deadlines surround \creature, who shakes.}
		\prl{\quad\creatureabbr Fie, get thee off me!}
		\prs{\deadlineone comes closer, breathing down \creature's neck. The \creature freezes.}
		\prl{\quad\creatureabbr I said no!}
		\prs{They strike at \deadlineone, then faint in fatigue. 
			\deadlineone staggers and retreats.
			All other \deadlines disappear into the distance. }
		\prs{}
		\prs{\phantomsection\addcontentsline{toc}{subsection}{Scene III}\hfil\textsc{Scene III.}---The forest, in \creature's dream.}
		\prs{Enter \hyperbard, with a lute.}
		 \prl{\quad\hyperbardabbr What beauty are these woods! In every tree}
		 \prl{Lives past enshrined and calling the observant.}
		 \prl{The devil? Angels lie in all these details.}
		 \prl{Look at the fragile bark, the fractal branching,}
		 \prl{The posture, parasites---And see the leaves!}
		 \prl{Colors, shapes, textures---all varieties.}
		 \prl{The fauna---beetles, rodents, insects, birds,}
		 \prl{Thriving together in their interaction.}
		 \prs{They strike a chord on their lute.}
		 \prl{\quad\hyperbardabbr From all there is, let there be data!}
		 \prl{As data points, we demarcate these trees}
		 \prl{And put them into known categories.}
		 \prs{They mark the selected trees with leaves of various shapes (Fig.~\hyperlink{fig:rawdata}{1}).}
		 \prl{\quad\hyperbardabbr Each tree is full of life, full of relations,}
		 \prl{To capture this, we need representations.}
		 \prs{They strike another chord. Enter \graph.}
		 \prl{\quad\graphabbr You called me, honor?}
		 \prl{\quad\hyperbardabbr \hspace*{9em}Will you, docile spirit,}
		 \prl{Transform these trees to yield discoveries?}
		 \prl{\quad\graphabbr Your honor, master, mistress, sure I can}
		 \prl{But there are many different transformations}
		 \prl{Among the flurry, which one do you choose?}
		 \prl{\quad\hyperbardabbr Why choose but one when there exist so many?}
		 \prl{How do we even know which one to pick?}
		 \prl{\mbox{\quad\graphabbr Sir, madam, with respect, your speech is madness!}}
		 \prl{Did you not call me to produce your truth?}
		 \prl{\mbox{\quad\hyperbardabbr What truth? Your transformations are but shadows}}
		 \prl{Of essence vested with complexity}
		 \prl{Cast on the narrow walls of our perception}
		 \prl{And varied as you shift and change your light.}
		 \prl{\mbox{\quad\graphabbr I hear your words but struggle with their meaning.}}
		 \prl{Which output do you want me to obtain?}
		 \prl{\quad\hyperbardabbr To every data point associate}
		 \prl{A set of transformations as its data.}
		 \prl{Such that in all our future inquiries}
		 \prl{We treat not only one but many shadows.}
		 \prl{Each partly blind, together they create}
		 \prl{A truer truth than commonly considered.}
		 \prl{\quad\graphabbr Your honor, as a practicality}
		 \prl{We can't enumerate exhaustively.}
		 \prl{Among the myriad possibilities}
		 \prl{You still will have to choose some transformations.}
		 \prl{\quad\hyperbardabbr Fair spirit, as an overarching goal,}
		 \prl{All our representations should be faithful.}
		 \prl{Among the transformations that you see,}
		 \prl{How do they differ systematically?}
		 \prs{Screaming heard. \hyperbard and \graph vanish.
		 	\creature wakes. 
	 	}
	\end{tabular}
\end{figure}






\clearpage 
\begin{figure}[h]\small
	\hspace*{-4em}\begin{tabular}[c]{rp{0.5\linewidth}|}
		&\hypertarget{fig:rawdata}{}
\tikzset{%
  every node/.append style = {%
    font = \scriptsize,
  },
}
\begin{tikzpicture}
  \begin{axis}[%
    axis x line* = bottom,
    axis y line* = left,
    tick align   = outside,
    legend pos   = south east,
    xlabel       = {Number of speaking characters},
    ylabel       = {Number of spoken lines},
    width        = 0.5\textwidth,
    height       = 0.5\textwidth,
  ]
    \addplot[%
      scatter, only marks,
      point meta      = explicit symbolic,
      scatter/classes = {%
        comedy={emerald, mark = *},
        tragedy={cardinal, mark = pentagon*},
        history={bleu,  mark = square*}%
      }
    ] table[x = n_characters, y = n_lines, meta index = 4, col sep = comma] {data/summary_statistics_raw.csv};

    \node[pin=0:{Romeo \& Juliet}] at (38, 3175) {};
    \node[pin=0 :{The Tempest}] at (19 , 2360) {};
    \node[pin=90 :{Richard III}] at (59, 3767) {};

    \legend{Comedy, Tragedy, History}
  \end{axis}
\end{tikzpicture}\\
		\stepcounter{figure}
		\pls{\emph{Figure~\thefigure: Number of spoken lines vs. number of speaking characters in the 37 plays by William Shakespeare. Each point corresponds to a play for which we provide 18 different (hyper)graph representations.}}
		\pls{}
		\pls{\phantomsection\addcontentsline{toc}{section}{Act II}\hfil\textsc{Act II.}}
		\pls{\phantomsection\addcontentsline{toc}{subsection}{Scene I}\hfil\textsc{Scene I.}---The Community. In the dining hall.}
		\pls{\professor, \seniorresearcher, and \colleague seated at a table.
			Enter \creature, carrying a tray.}
		\pll{\quad\colleagueabbr Hey fellow, please come join us, have a seat!}
		\pls{\creature, jolted from their thoughts, obeys with reluctan\-ce.}
		\pll{\quad\seniorresearcherabbr They told me you submitted, so, good cheer!}
		\pll{\quad\colleagueabbr Next time, though, try to not scare \secretary.}
		\pll{\quad\professorabbr Now fate lies with the review gods, almighty}
		\pll{And they select not just for quality.}
		\pll{Regardless of their upcoming decision,}
		\pll{You'll get this published, well, eventually.}
		\pln{\quad\creatureabbr That's comforting.}
		\pll{\quad\colleagueabbr \hspace*{7.5em}Well, it is how things go.}
		\pll{\quad\professorabbr My admin work is calling.}
		\pln{\quad\seniorresearcherabbr \hspace*{10em}And mine, too!}
		\pls{Exeunt \professor and \seniorresearcher.}
		\pls{Awkward silence.}
		\pll{\quad\creatureabbr May I ask you something?
			Here in the Community, how do you get your data? You hardly go outside...
		}\addtocounter{playlinenumber}{1}
		\pll{\quad\colleagueabbr What do you mean? We grab it from the shelves.}
		\pll{There's shelves for almost every data type.}
		\pll{For graphs, e.g., there's OGB \cite{hu2020ogb}, and SNAP \cite{leskovec2016snap},}
		\pll{KONECT \cite{kunegis2013konect}, and TUD \cite{morris2020}, and Netzschleuder \cite{peixoto2020netzschleuder},}
		\pll{And finally, Network Repository \cite{rossi2015network}.}
		\pll{\quad\creatureabbr Hold on, you are confusing me.
			How do the graph shelves get their data, then? 
		}\addtocounter{playlinenumber}{1}
		\pll{\quad\colleagueabbr You really ask the weirdest things. I guess}
		\pll{They send some hunter-gatherers to catch}
		\pll{Or pick the graphs they find out in the wild.}
		\pll{\quad\creatureabbr You make it sound like graphs exist, for real. 
			But are they not defined by their observers?\hypertarget{inquisition}{}
		}\addtocounter{playlinenumber}{1}
		\pll{\quad\colleagueabbr Who are you? Not the Spanish Inquisition?}
		\pll{All graphs have nodes and edges, that's what matters.}
		\pll{Sometimes they come with weights or attributes.}
		\pll{Semantics---God, who cares?---graphs are abstractions,}
		\pll{And abstract data is our working truth.}
		\pls{Exeunt.}
	\end{tabular}
	~
	\begin{tabular}[c]{p{0.5\linewidth}r}
		\prs{\phantomsection\addcontentsline{toc}{subsection}{Scene II}\hfil\textsc{Scene II.}---\creature's office.}
		\prs{In a corner, on the floor, \creature, in contemplation.}
		\prl{\quad\creatureabbr What canny creatures met my febrile mind.
			That friendly faun, the gentle spirit, 
			exchanging such profound considerations. 
			I wish I could have stayed a little longer---instead, I'm left to draw my own conclusions. 
			What graph shadows could I create
			by shining different lights on what there is?
			It seems the sensible depends on the semantics.
		}\addtocounter{playlinenumber}{6}
		\prs{They close their eyes, following their thoughts.}
		\prl{\quad\creatureabbr  
			When we transform reality to math,}
		\prl{
			Graphs are but outputs, 
			in---phenomena.}
		\prl{%
			The myriad transformations that we see,} 
		\prl{How do they differ systematically?}
		\prl{For now, we shall distinguish three dimensions.}
		\prl{First, our \emph{semantic mapping}---Nodes and edges:}
		\prl{What types of entities do we assign?} 
		\prl{Second, our \emph{granularity}---What are} 
		\prl{Our modeling units for semantic mapping?}
		\prl{And third, our \emph{expressivity}: What more}
		\prl{Do we attach to all our modeling units?}
		\prl{Directions, weights, and multiplicities,}
		\prl{Or attributes and cardinalities...} 
		\prl{What universe! \emph{Haec facta, fiant data.}}
		\prs{Tracing coordinate axes with their fingers, they sigh.}
		\prl{\quad\creatureabbr All these distinctions, it appears, are known in the Community \cite{torres2021representations}. 
			And yet, the knowledge seldom heeded---graph data shelves are filled with all these captive singular truths.
			We hardly hold what that free faun foresaw: 
			For every data point, a set of transformations as its data. 
			I wonder why.
		}\addtocounter{playlinenumber}{4}
		\prs{Exit.}
		\prs{}
		\prs{\phantomsection\addcontentsline{toc}{subsection}{Scene III}\hfil\textsc{Scene III.}---\colleague's office.}
		\prs{\colleague, trimming a bonsai with scissors. 
		}
		\prl{\quad\colleagueabbr Alas, they really want documentation?} 
		\prs{\creature steps into the door frame, unnoticed.\hypertarget{alltheworld}{}}
		\prl{\quad\colleagueabbr A datasheet \cite{gebru2021datasheets}? Well---all the world is data,}
		\prl{And all we care for merely data points;}
		\prl{They get created, updated, deleted,}
		\prl{And every data point plays many parts,}
		\prl{Its fate being seven stages. First, \emph{motivation}}
		\prl{Defining purpose or specific tasks.}
		\prl{Then \emph{composition}, sketching the raw data}
		\prl{And telling people where it was obtained,}
		\prl{If anything's amiss. And then \emph{collection},}
		\prl{How did we get each single data point,}
		\prl{And what else did we check. Then \emph{preprocessing},}
		\prl{Full of strange quirks and idiosyncrasies,}
		\prl{But made that it looks principled. Then \emph{uses},}
		\prl{What all things did we do, what could have been,}
		\prl{And what should not be done. Then \emph{distribution},}
		\prl{If, when, and how will we make data public,}
		\prl{Restrictions by third parties, if imposed,}
		\prl{And also all the laws. Last stage of all,}
		\prl{That ends this template documentary,}
		\prl{Is \emph{maintenance} and hosting and support,}
		\prl{Sans updates, sans errata, sans comment.}
		\prs{\creature retires, flabbergasted.}
		\prs{\colleague stashes the stunted bonsai into a shelf.}
		\prs{Exit.}
	\end{tabular}
\end{figure}

\clearpage

\begin{figure}[h]\small
	\hspace*{-4.1em}\begin{tabular}[t]{rp{0.5\linewidth}|}
		\pls{\phantomsection\addcontentsline{toc}{subsection}{Scene IV}\hfil\textsc{Scene IV.}---\creature's office.}
		\pls{Enter \creature, restless.}
		\pll{\quad\creatureabbr This stream of observations leaves me drowning in confusion. 
			If \emph{Is} is not what \emph{Ought}, how can \emph{Is} be?
			What is this thing they call Community?
			Am I misguided, am I wrong---to doubt that I belong?\hypertarget{faust}{}
		}\addtocounter{playlinenumber}{3}
		\pll{Two souls, alas, are dwelling in my breast,}
		\pll{And each one seeks to rule without the other.}
		\pll{The one a falcon, fierce and fighting fetters,}
		\pll{That's dreaming of faun's forest, flying free,}
		\pll{The other a caged chary canary,}
		\pll{That calmly, coyly, cheerfully chants chatters.}
		\pls{They open the window and balance on the window sill.}
		\pll{\quad\creatureabbr Should there be spirits roaming through the air,}
		\pll{I beg they lift the spell of my despair.}
		\pls{They jump.}
		\pls{}
		\pls{\phantomsection\addcontentsline{toc}{subsection}{Scene V}\hfil\textsc{Scene V.}---The forest.}
		\pls{\graph tending to a mat of moss. On the mat, \creature, somnolent. Enter \hyperbard.}
		\pll{\quad\hyperbardabbr So few return once captured by Its magic!}
		\pll{\quad\graphabbr Playing that dream was worth it, after all.}
		\pll{\quad\creatureabbr Is this a dream no more? Do you exist?}
		\pll{\quad\hyperbardabbr Depends on your philosophy. But see,}
		\pll{My \graph says you have interesting ideas.}
		\pll{So tell me, how would \emph{you} transform these trees}
		\pll{To bear the fruit of new discoveries?}
		\pll{\quad\creatureabbr Did you not eavesdrop on my ruminations,}
		\pll{Distinguishing between those three dimensions?}
		\pll{Semantic mapping, granularity,}
		\pll{And expressivity---put abstractly?}
		\pll{\quad\hyperbardabbr I heard, but what does it all mean in practice?}
		\pll{\quad\creatureabbr Let's walk through an example. Take this tree:}
		\pll{The Tragedy of R. and J.---a play.}
		\pll{When modeled \emph{Les Mis\'erables}-y \cite{knuth1993stanford}, the nodes}
		\pll{Are characters, and edges---co-occurrence.}
		\pll{That's one semantic mapping, hold this fixed.}
		\pll{Then, as to granularity, we ask}
		\pll{What unit should determine co-occurrence?}
		\pll{The first---most common---option is: a scene.}
		\pll{And here, much modeling ends, unfortunately:}
		\pll{Max simple graphs, min expressivity.}
		\pll{\quad\hyperbardabbr But does this not reveal essential structure?}
		\pll{\quad\creatureabbr It smudges all the details, Fig.~\ref{fig:representations:ce-scene:a}!}
		\pll{Do the play's namesake heroes co-occur}
		\pll{No more than Montague and Capulet?}
		\pll{\quad\hyperbardabbr So should we count-weight edges, Fig.~\ref{fig:representations:ce-scene:b}?}
		\pll{\quad\creatureabbr Or introduce edge multiplicity.}
		\pll{The multigraph perspective would allow us}
		\pll{To treat---Fig.~\ref{fig:representations:ce-scene:c}---%
			co-occurrence weights.}
		\pll{In our setting, this could, e.g., mean}
		\pll{The count of spoken lines in every scene.}
		\pll{But that is basic expressivity---}
		\pll{We yet have to treat granularity.}
		\pll{To illustrate, in Fig.~\ref{fig:representations:ce:a}, we draw}
		\pll{The co-occurrence only for Act III.}
		\pll{The Capulets and Romeo appear}
		\pll{To interact too much---this sparks suspicion.}
		\pll{\quad\hyperbardabbr You mean we're introducing information?}
		\pll{\quad\creatureabbr And hiding what there really is to see!}
		\pll{The scene is far too coarse a modeling unit,}
		\pll{Quite often is there movement in between.}
		\pll{We must keep track of entries and of exits}
		\pll{To capture interactions faithfully.}
		\pll{Each part confined by any two such changes,}
		\pll{A \emph{stage group}, separately defines an edge.}
	\end{tabular}~
	\begin{tabular}[t]{p{0.5\linewidth}r}
		\prl{Accounting now for expressivity,}
		\prl{These edges may be binary or multi,}
		\prl{Or weighted by lines spoken, Fig.~\ref{fig:representations:ce:b}.}
		\prl{The outcome, evident from  Fig.~\ref{fig:representations:ce:c},}
		\prl{Is far from what we had initially.}
		\prl{Thus, even for just one semantic mapping,}
		\prl{And R. and J. as a specific case:}
		\prl{We see at least six decent transformations,}
		\prl{Statistics differing tremendously.}
		\prl{\quad\hyperbardabbr So is this all?}
		\prn{\quad\creatureabbr \hspace*{6em}Oh, that is but the start!}
		\prl{Thus far, we've had just characters as nodes.}
		\prl{One possible complaint with this approach}
		\prl{Is that it gives us artificial cliques.}
		\prl{Instead, we could in our semantic mapping}
		\prl{Consider also parts of plays as nodes,}
		\prl{Transforming plays into bipartite graphs,}
		\prl{Whose edges signal character occurrence.}
		\prl{Then granularity, Fig.~\ref{fig:representations:se:a}--b,}
		\prl{Concerns the nodes, but sometimes also edges.}
		\prl{In terms of expressivity, we could}
		\prl{Again attend to weights, and represent}
		\prl{Directionality, see Fig.~\ref{fig:representations:se:c},}
		\prl{With greater ease than in the one-mode case---}
		\prl{To model single \emph{speech acts}, too, as edges.}
		\prl{\quad\hyperbardabbr Now, that is quite a lot---so are you finished?}
		\prl{\quad\creatureabbr Respectfully, the best is yet to come!}
		\prl{Conceptually, all I have just described}
		\prl{Can be derived from a more general model.}
		\prl{All graphs, regarding expressivity}
		\prl{Force `$\in\{1,2\}$' on cardinality}
		\prl{Of edges---}
		\prn{\quad\hyperbardabbr \hspace*{2em}Marvelous mathematically!}
		\prl{\quad\creatureabbr But artificial, thinking critically.}
		\prl{The interactions in your vivid woods---}
		\prl{How many of them are bilateral?}
		\prl{This common cardinality constraint:}
		\prl{Let's do away with it!}
		\prn{\quad\hyperbardabbr\hspace*{6em}Then what remains?}
		\prl{\quad\creatureabbr A set system---a \emph{hypergraph}, they say \cite{berge1984hypergraphs},}
		\prl{We visualize its power in Fig.~\ref{fig:representations:hg}.}
		\prl{Confusingly: All graphs are hypergraphs}
		\prl{But not vice versa.}
		\prn{\quad\hyperbardabbr\hspace*{4.5em} Do we need this, \graph?}
		\prl{\quad\graphabbr Well, some found hypergraphs to be quite handy}
		\prl{To capture higher-order interactions \cite{bai2021hypergraph,battiston2021physics,aksoy2020hypernetwork}.}
		\prl{They certainly are more intuitive}
		\prl{Than making cliques of higher arities,}
		\prl{Or else treating relations, too, as nodes.}
		\prl{\quad\creatureabbr We can go far with \emph{graphs} but don't know yet}
		\prl{Just how much further we can get with \emph{hyper}.}
		\prl{Observe the beauty in these hypergraphs:}
		\prl{They readily entail \emph{all} transformations!}
		\prl{From their perspective, what first we discussed}
		\prl{Are \emph{clique expansions}, and our next ideas}
		\prl{Are known as \emph{star expansions} \cite{srinivasan2021learning}---see, in sum,}
		\prl{Fig.~\ref{fig:all}, and our proposals in Tab.~\ref{tab:representations}.}
		\prl{\quad\hyperbardabbr Things hyper, in their generality,}
		\prl{They seem to suit my woods quite naturally.}
		\prl{\quad\graphabbr But sovereign, as a practicality,}
		\prl{There's hardly any software letting us}
		\prl{Compute with hypergraphs conveniently!}
		\prl{\quad\hyperbardabbr \emph{and} \creatureabbr \emph{[in synchrony]} Who are you, the Community?}
		\prn{\quad\graphabbr\hspace*{2.5em} I'm sorry.}
		\prs{Exeunt.}
	\end{tabular}
\end{figure}

\clearpage
\begin{figure}[t]
	\centering
	\begin{subfigure}{0.33\linewidth}
		\includegraphics[height=4cm]{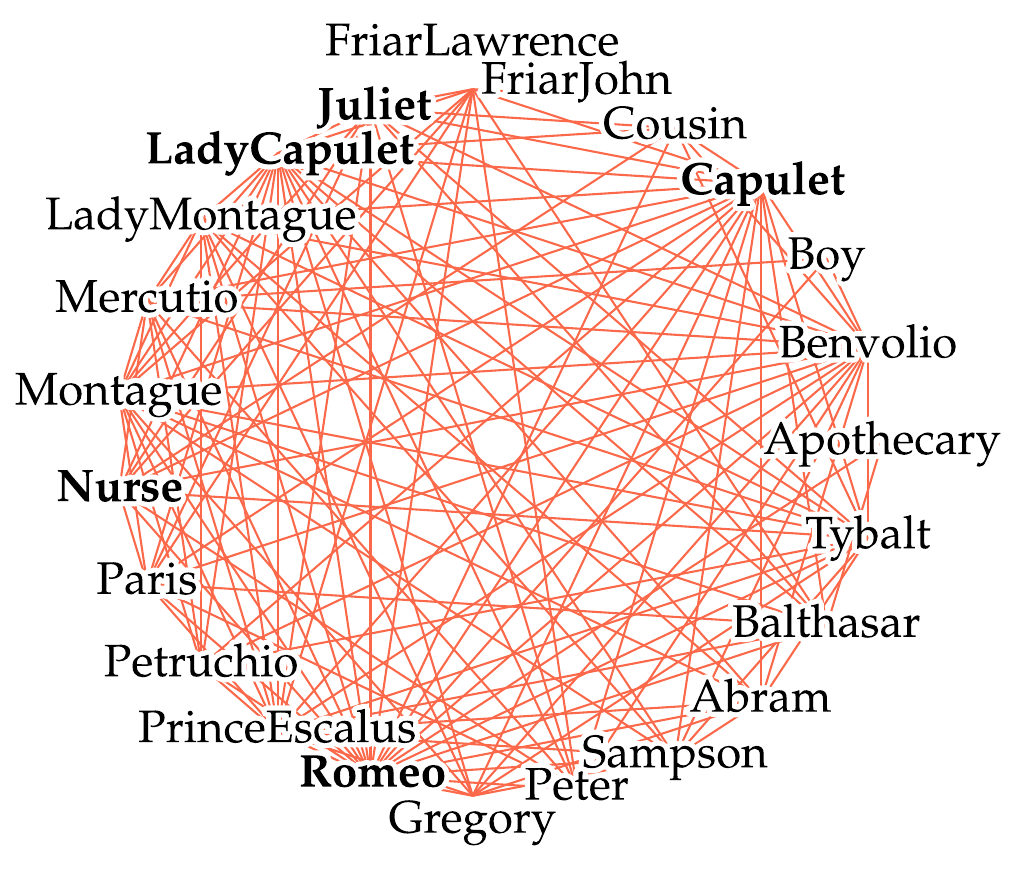}~%
		\subcaption{ce-scene-b}\label{fig:representations:ce-scene:a}
	\end{subfigure}~
	\begin{subfigure}{0.33\linewidth}
		\includegraphics[height=4cm]{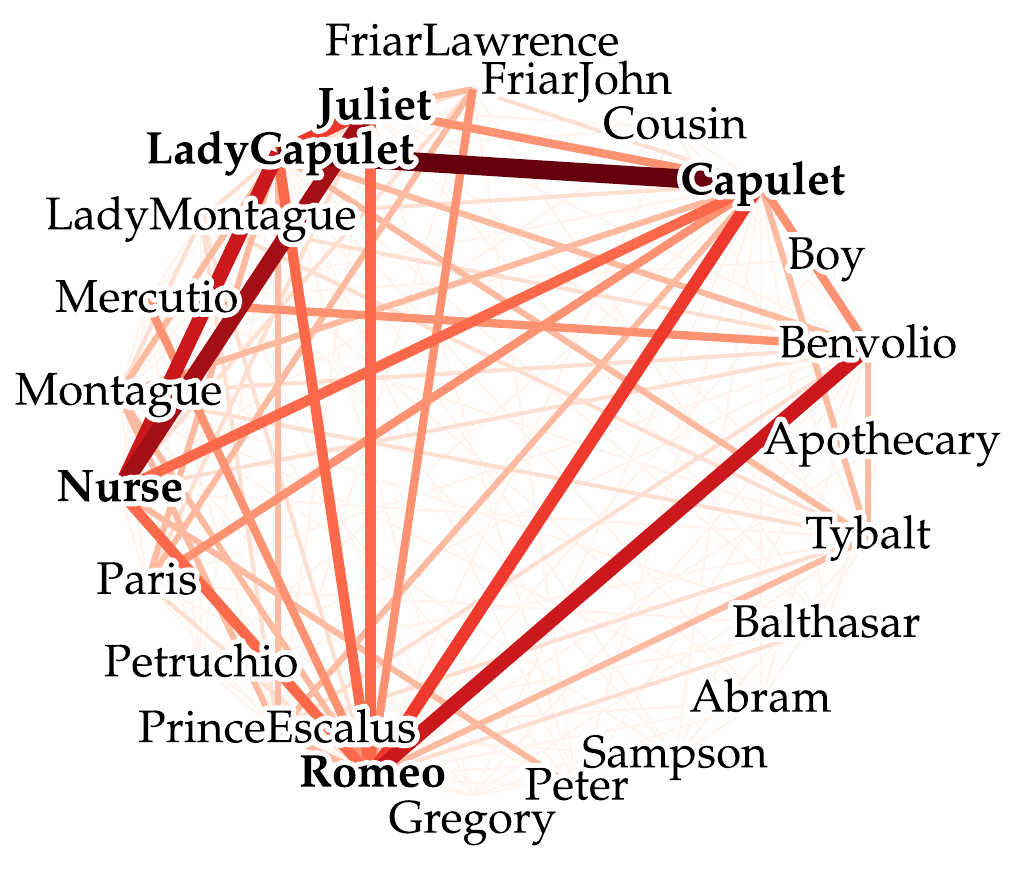}~%
		\subcaption{ce-scene-mb $\sim$ ce-scene-w}\label{fig:representations:ce-scene:b}
	\end{subfigure}~
	\begin{subfigure}{0.33\linewidth}
		\centering
		\includegraphics[height=4cm]{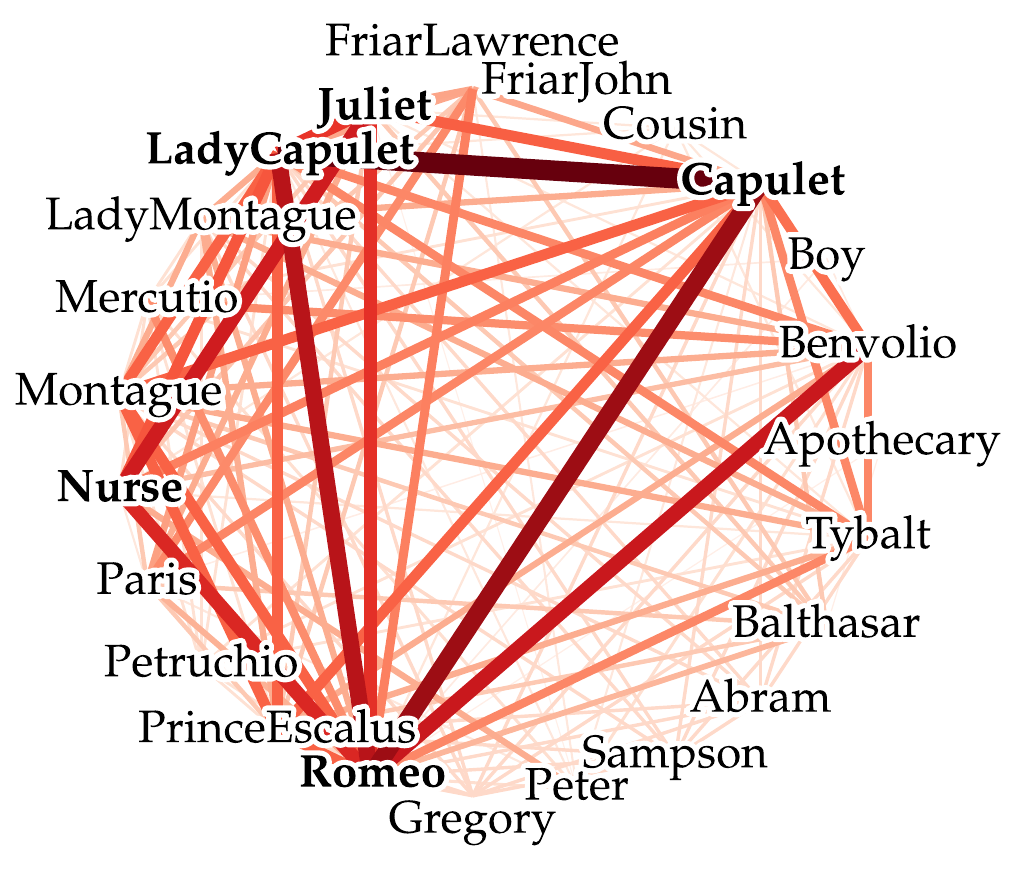}~%
		\subcaption{ce-scene-mw}\label{fig:representations:ce-scene:c}
	\end{subfigure}%
	\caption{%
		Relationships between the named characters in \randj when modeled as binary~(a), 
		count-weighted~(b), 
		and line-weighted~(c) co-occurrence networks, resolved at the scene level, 
		where we highlight the protagonists appearing in Act~III, Scene~V. 
		The binary representation is a classic hairball, 
		while the count-weighted representation and the line-weighted representation provide more nuance. 
		In (c), the strikingly strong connection between Romeo and Capulet is partly due to Act III, Scene V, 
		where both characters appear but \emph{do not meet} on stage.
	}\label{fig:representations:ce-scene}
\end{figure}
\begin{figure}[t]
	\centering
	\vspace*{-12pt}\begin{subfigure}{0.33\linewidth}
		\centering
		\includegraphics[height=4cm]{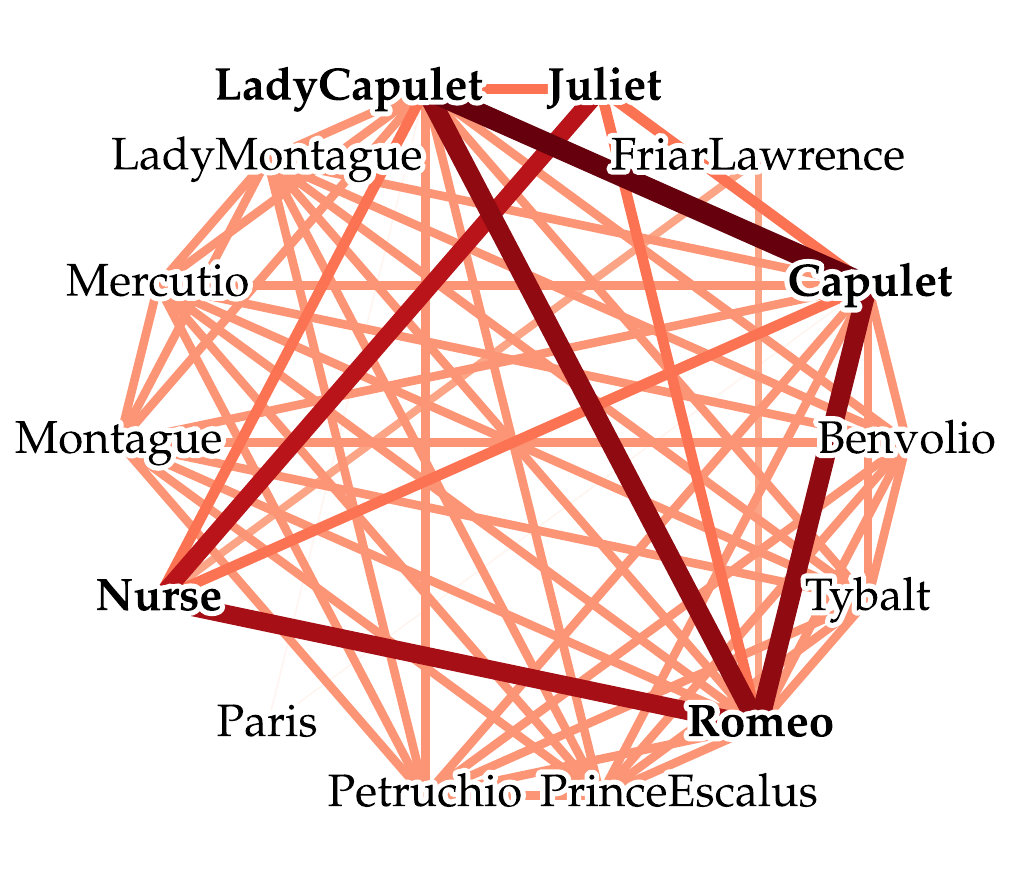}
		\subcaption{ce-scene-mw}\label{fig:representations:ce:a}
	\end{subfigure}~
	\begin{subfigure}{0.33\linewidth}
		\centering
		\includegraphics[height=4cm]{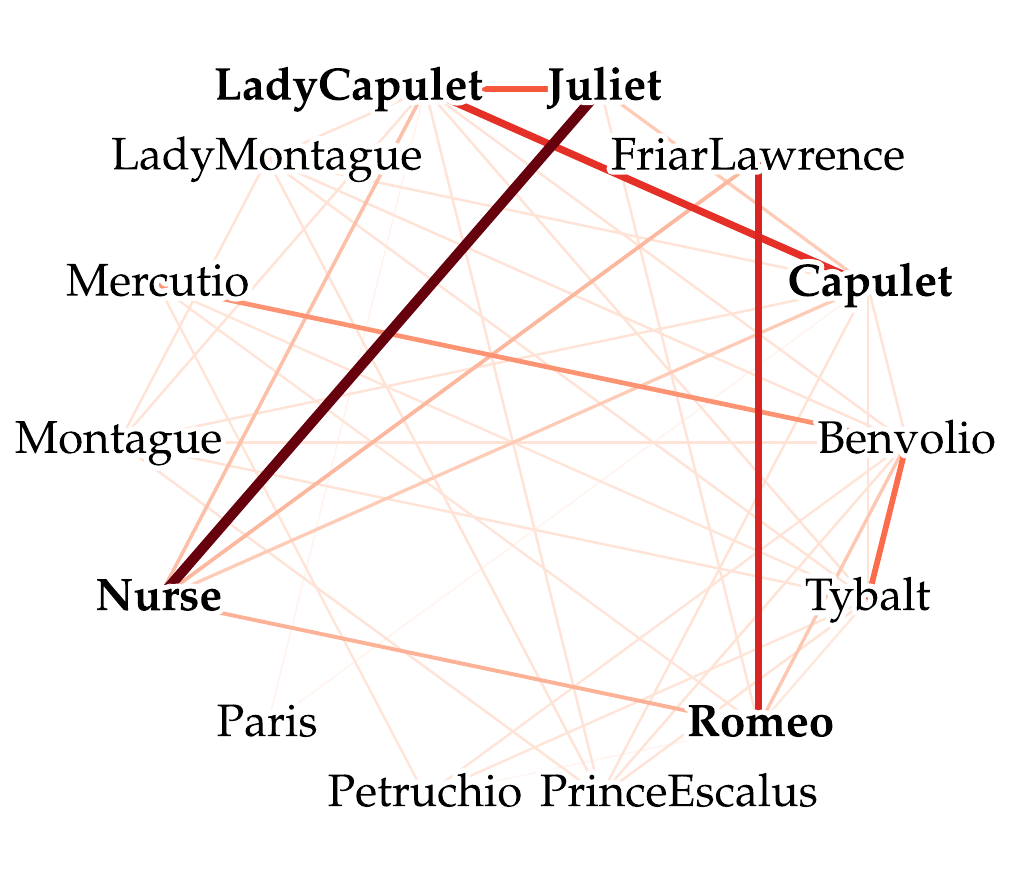}
		\subcaption{ce-group-mw}\label{fig:representations:ce:b}
	\end{subfigure}~
	\begin{subfigure}{0.33\linewidth}
		\centering
		\includegraphics[height=4cm]{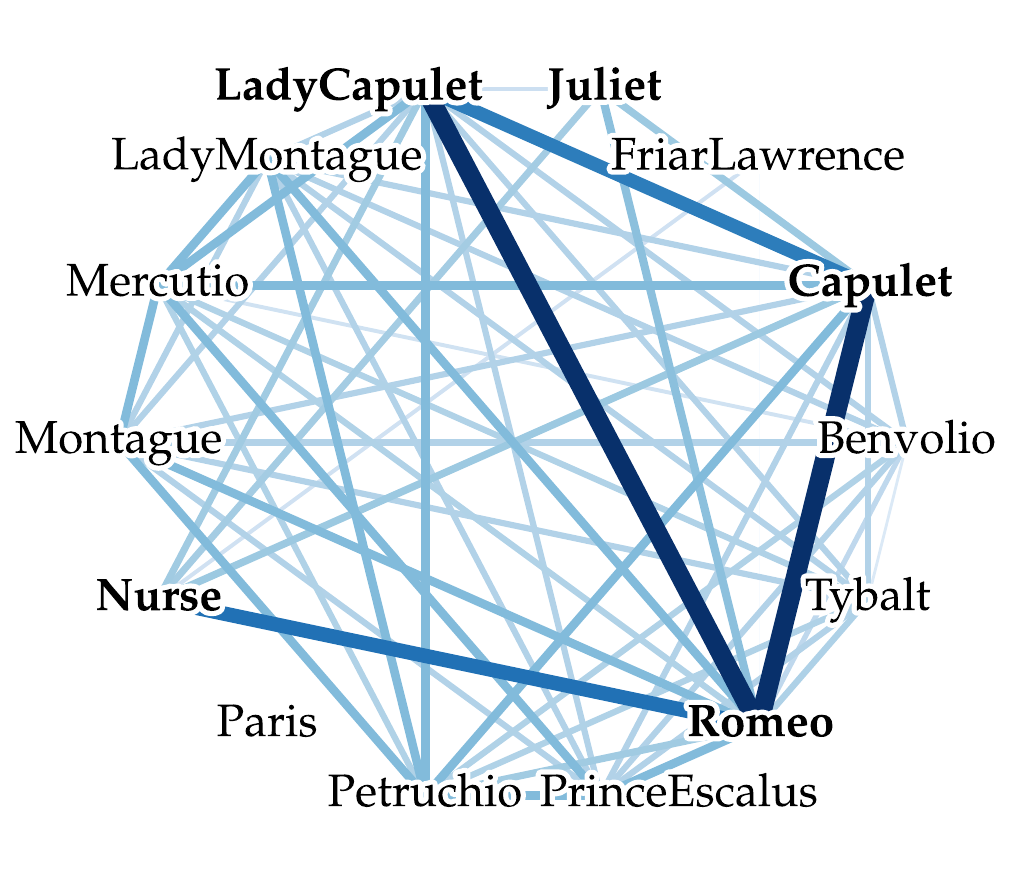}
		\subcaption{(ce-scene-mw) $-$ (ce-group-mw)}\label{fig:representations:ce:c}
	\end{subfigure}
	\caption{Line-weighted co-occurrence network of the named characters in Act~III of \randj, 
		resolved at the scene level (a) and 
		at the stage group level (b), 
		as well as the difference network between the two (c),
		where we highlight the protagonists appearing in Act~III, Scene~V.  
		The coarse-grained representation overestimates the co-occurrence between Juliet's parents (Capulet and Lady Capulet) and Romeo (a and c),
		while the fine-grained representation emphasizes Juliet's bond with the Nurse and Romeo's interaction with Friar Lawrence (b).
	}\label{fig:representations:ce}
\end{figure}
\begin{figure}[t]
	\centering
	\vspace*{-12pt}\begin{subfigure}{0.28\linewidth}
		\centering
		\includegraphics[height=4cm]{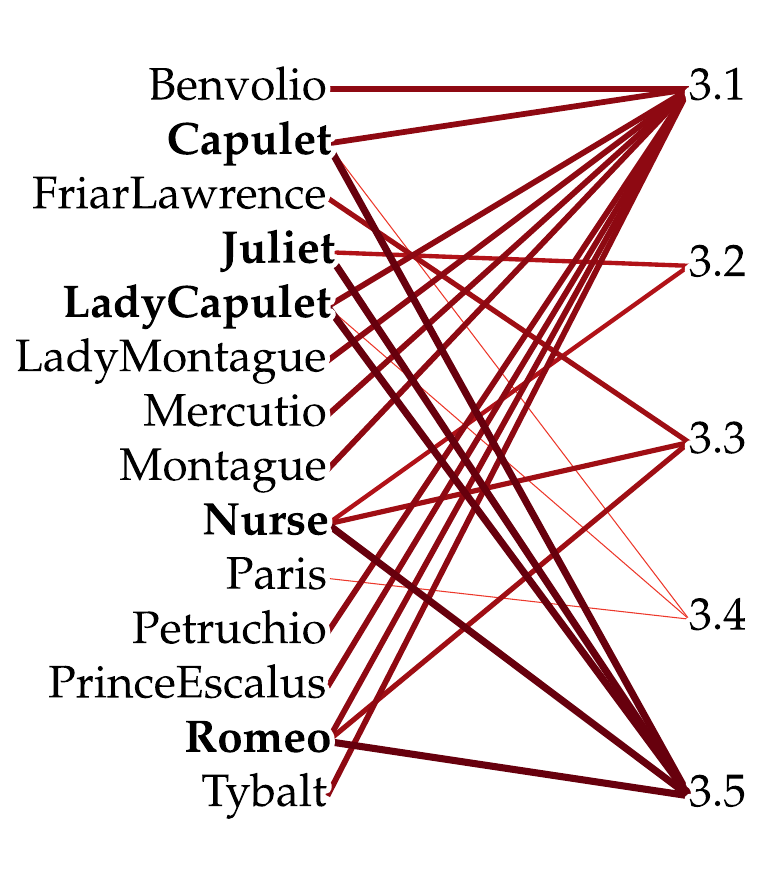}~%
	\end{subfigure}~
	\begin{subfigure}{0.28\linewidth}
		\centering
		\hspace*{2em}\includegraphics[height=4cm]{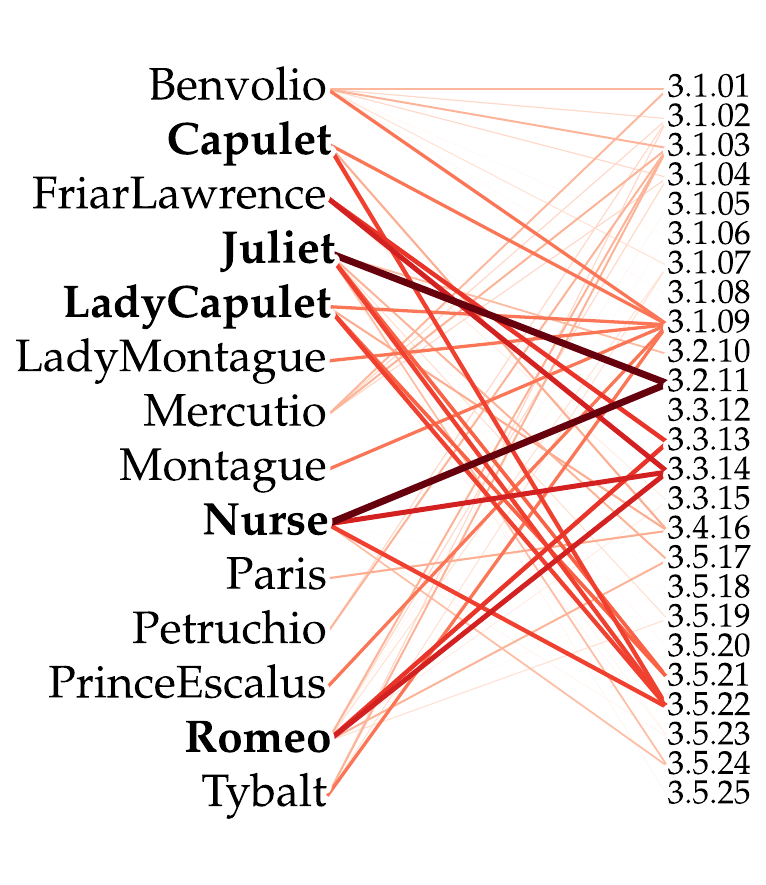}~%
	\end{subfigure}~
	\begin{subfigure}{0.44\linewidth}
		\centering
		\includegraphics[height=4cm]{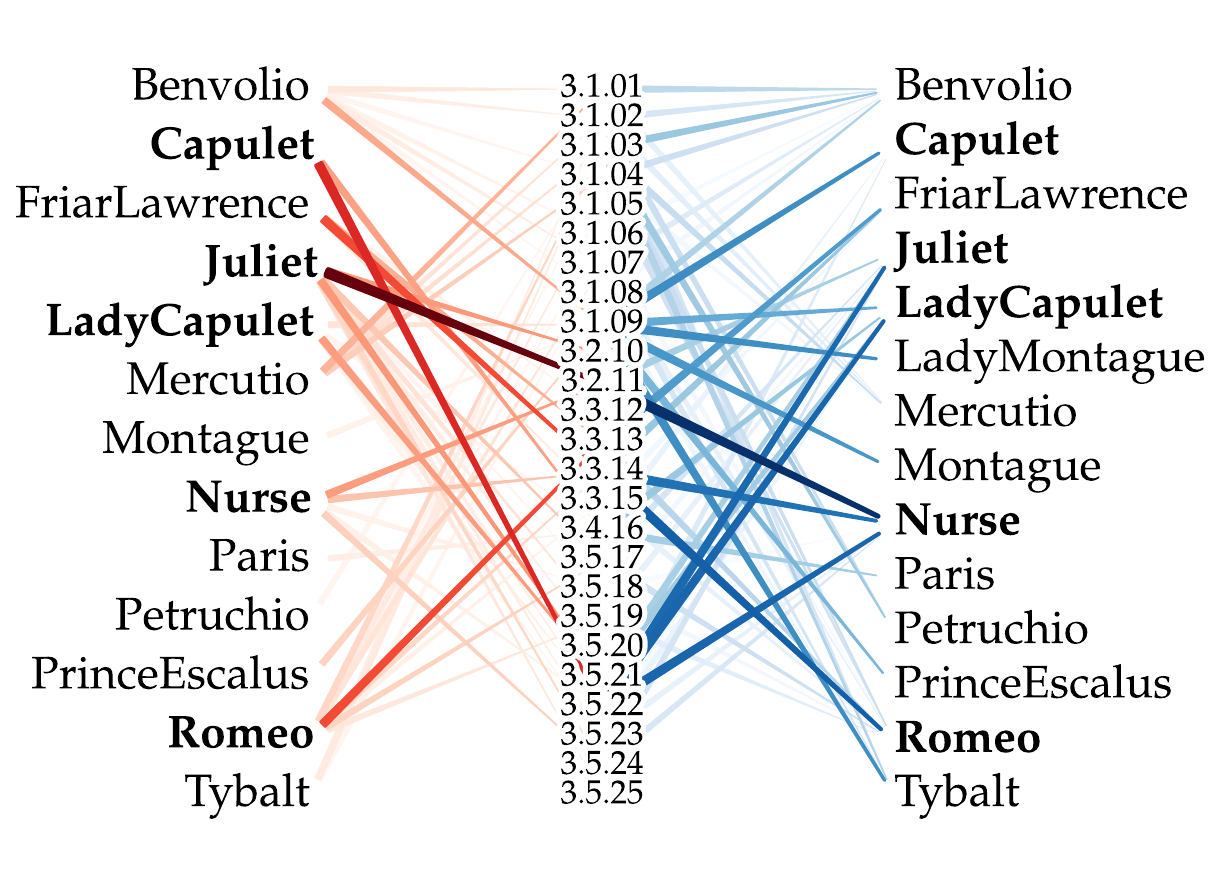}
	\end{subfigure}
\begin{subfigure}{0.33\linewidth}
	\subcaption{se-scene-w}\label{fig:representations:se:a}
\end{subfigure}~
\begin{subfigure}{0.33\linewidth}
	\subcaption{se-group-w}\label{fig:representations:se:b}
\end{subfigure}~
\begin{subfigure}{0.33\linewidth}
\subcaption{se-speech-wd}\label{fig:representations:se:c}
\end{subfigure}
	\caption{Weighted bipartite graph of named character occurrences in Act~III of \randj, 
		resolved at the scene level (a) and 
		at the stage group level (b), as well as the directed weighted bipartite graph resolved at the speech act level, with character nodes split up into speakers and listeners for visual clarity (c),
		where we highlight the protagonists appearing in Act~III, Scene~V.  
		While the coarse-grained representation  overestimates Romeo's role in Act~III, Scene V (a), 
		the finer-grained representation again highlights Juliet's bond with the Nurse (b), 
		and the directed representation reveals the hierarchical structure of their communication (c). 
	}\label{fig:representations:se}
\end{figure}
\clearpage
\begin{landscape}
\begin{figure}[t]
		\centering
		\vspace*{-1.5cm}\begin{subfigure}[b]{0.25\linewidth}
			\centering
			\begin{tabular}{l}
				$\mid$$\rightarrow$$A; A*$$\mid$$\rightarrow$$B; A*$$\mid $\\
				$\rightarrow$ $C;$$B*; A$$\rightarrow$$\mid$ $C*; B$$\rightarrow$$\mid$\\ 
				$C*$$\mid$$\rightarrow$$D; D*$$\mid$$\rightarrow$$A, B, E;$\\
				$A*; A, B, C, D, E \rightarrow \mid$\\
			\end{tabular}
		\subcaption{Toy drama}
		\end{subfigure}~
		\begin{subfigure}[b]{0.25\linewidth}
			\centering
			\includegraphics[height=2cm]{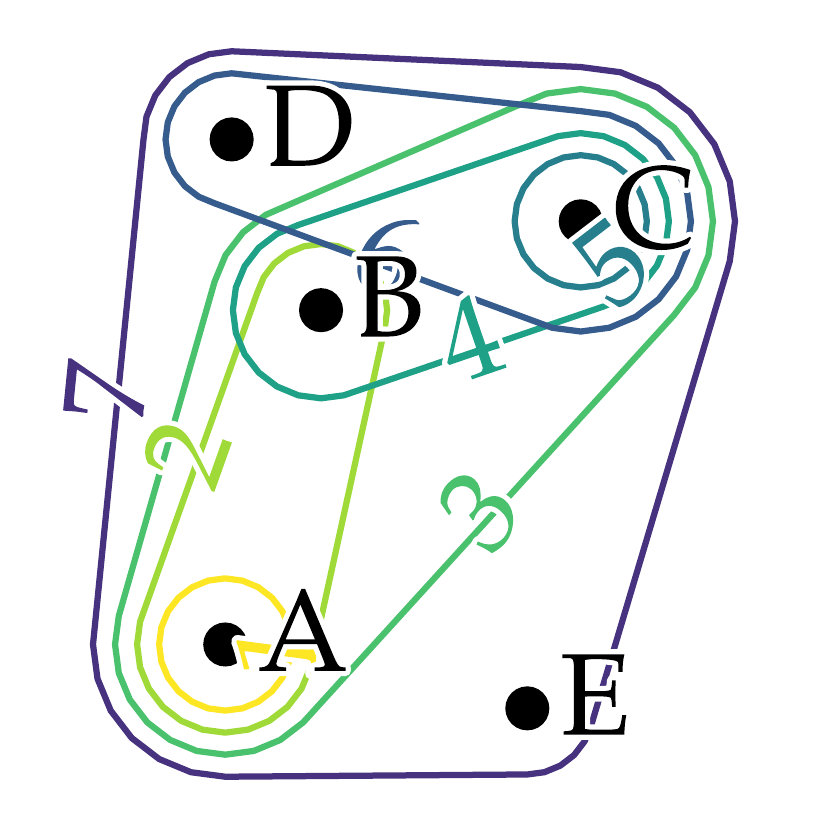}
			\subcaption{Hypergraph ($n=5, m=7$)}
		\end{subfigure}~
		\begin{subfigure}[b]{0.25\linewidth}
			\centering
			\includegraphics[height=2cm]{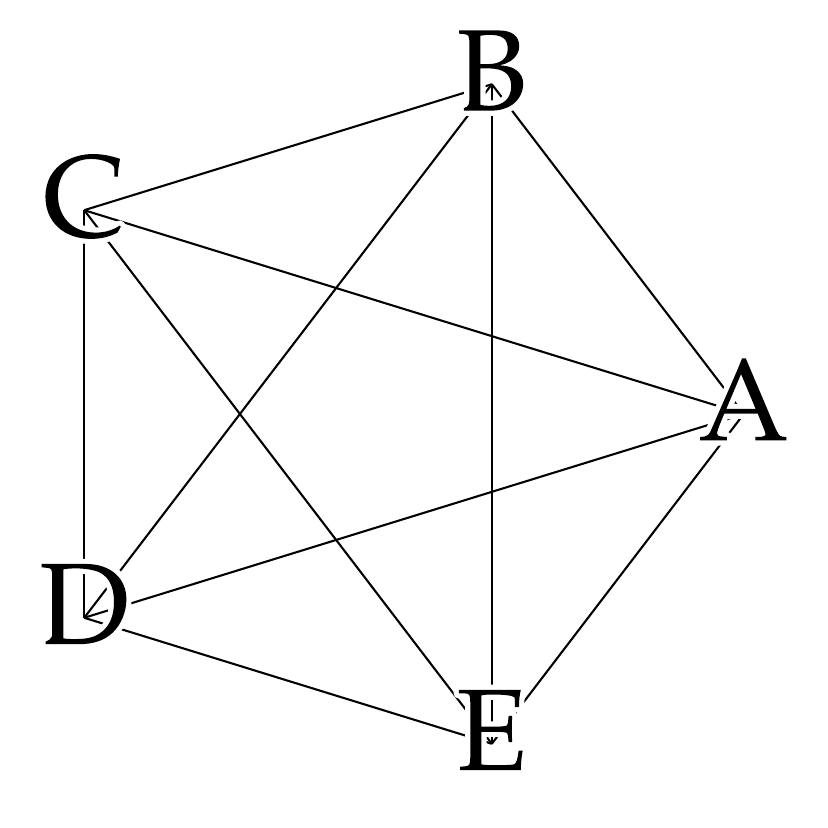}
			\subcaption{Clique expansion ($n=5, m=10$)}
		\end{subfigure}~
		\begin{subfigure}[b]{0.25\linewidth}
			\centering
			\includegraphics[height=2cm]{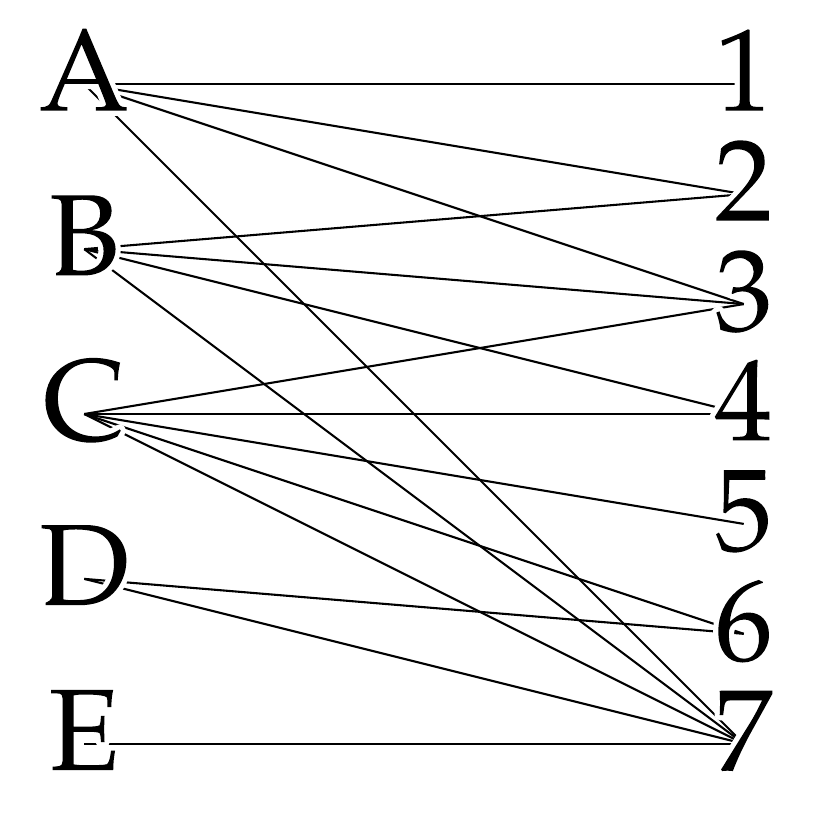}
			\subcaption{Star expansion ($n=12, m=16$)}
		\end{subfigure}
		\caption{Relationship between hypergraphs, clique expansions, and star expansions, illustrated for a toy drama. 
			In the toy drama, characters are capital letters, $\rightarrow$$X$ denotes entry, $X$$\rightarrow$ denotes exit, $*$ denotes speech, $\mid$ marks scene boundaries, $;$ marks activity boundaries, and $,$ indicates several characters acting together.
		}\label{fig:all}
	\end{figure}
\begin{table}[t]\small
	\centering
	\caption{Overview of relational data representations provided with \ourdata for each play attributed to William Shakespeare, based on the TEI simple-encoded XMLs provided by \folgertexts \cite{folger2022}. 
		Unidirectional arrows indicate assignment; bidirectional arrows indicate bijection.
		We highlight the transformations most commonly used in the literature.
	}\label{tab:representations}\vspace{6pt}
	
	\begin{tabular}{p{0.1\linewidth}p{0.335\linewidth}p{0.165\linewidth}p{0.32\linewidth}}
		\toprule
		\bfseries Representation&\bfseries Semantic Mapping&\bfseries Granularity&\bfseries Expressivity\\\midrule
		\textbf{ce-scene-b}&\multirow{6}{\linewidth}{\hspace*{-1em}$\left.\begin{array}{l}
				\\
				\\
				\\[3pt]
				\\
				\\
				\\
			\end{array}\right\rbrace$}&\multirow{3}{\linewidth}{\hspace*{-1em}$\left.\begin{array}{l}
			\\
			\\
			\\[-3pt]
		\end{array}\right\rbrace$ Edges $\leftrightarrow$ Scenes}&~---\\
		\textbf{ce-scene-mb}&&&Edge order\\
		ce-scene-mw&\hspace*{1.5em}Nodes $\leftarrow$ Characters &&Edge order, edge weights\\[3pt]
		ce-group-b&\hspace*{1.5em}Edges $\leftarrow$ Co-occurrence&\multirow{3}{\linewidth}{\hspace*{-1em}$\left.\begin{array}{l}
				\\
				\\
				\\[-3pt]
			\end{array}\right\rbrace$ Edges $\leftrightarrow$ Stage groups}&~---\\
		ce-group-mb&&&Edge order\\
		ce-group-mw&&&Edge order, edge weights\\
		\midrule
		se-scene-b&\multirow{4}{\linewidth}{\hspace*{-1em}$\left.\begin{array}{l}
				\\
				\\[3pt]
				\\
				\\
			\end{array}\right\rbrace$ Edges $\leftarrow$ Occurrence}~ \multirow{6}{*}{\hspace*{-12.25em}$\left.\begin{array}{l}
			\\
			\\[3pt]
			\\
			\\[3pt]
			\\
			\\
		\end{array}\right\rbrace$}   
		&\hspace*{-1em}\multirow{2}{*}{$\left.\begin{array}{l}
				\\
				\\[-3pt]
			\end{array}\right\rbrace$ Nodes (2) $\leftrightarrow$ Scenes}&Partial node and edge order\\
		se-scene-w&\hspace*{1.5em}&&Partial node and edge order; edge weights\\[3pt]
		se-group-b&\hspace*{14.75em}Nodes (1) $\leftarrow$ Characters&\hspace*{-1em}\multirow{2}{*}{$\left.\begin{array}{l}
				\\
				\\[-3pt]
			\end{array}\right\rbrace$ Nodes (2) $\leftrightarrow$ Stage groups}&Partial node and edge order\\
		se-group-w&\hspace*{14.75em}Nodes (2) $\leftarrow$ Play parts&&Partial node and edge order; edge weights\\[3pt]
		se-speech-wd&\multirow{2}{\linewidth}{\hspace*{-1em}$\left.\begin{array}{l}
				\\
				\\[-3pt]
			\end{array}\right\rbrace$~~Edges $\leftarrow$ Information flow}  
		&\hspace*{-1em}\multirow{2}{*}{$\left.\begin{array}{l}
				\\
				\\[-3pt]
			\end{array}\right\rbrace$}~Nodes (2) $\leftrightarrow$ Stage groups&Partial node order; edge weights, edge directions\\
		\mbox{se-speech-mwd}&&\quad\thinspace Edges \hspace*{1.525em}$\leftrightarrow$ Speech acts&\mbox{Partial node and edge order; edge weights, edge directions}\\
		\midrule
		hg-scene-mb&\multirow{4}{\linewidth}{\hspace*{-1em}$\left.\begin{array}{l}
				\\
				\\[3pt]
				\\
				\\
			\end{array}\right\rbrace$ Edges $\leftarrow$ Co-occurrence}~ \multirow{6}{*}{\hspace*{-12.25em}$\left.\begin{array}{l}
			\\
			\\[3pt]
			\\
			\\[3pt]
			\\
			\\
		\end{array}\right\rbrace$ Nodes $\leftarrow$ Characters}  &\hspace*{-1em}\multirow{2}{*}{$\left.\begin{array}{l}
				\\
				\\[-3pt]
			\end{array}\right\rbrace$ Edges $\leftrightarrow$ Scenes}&Edge order\\
		hg-scene-mw&&&Edge order, edge weights; edge-specific node weights\\[3pt]
		hg-group-mb&&\hspace*{-1em}\multirow{2}{*}{$\left.\begin{array}{l}
				\\
				\\[-3pt]
			\end{array}\right\rbrace$ Edges $\leftrightarrow$ Stage groups}&Edge order\\
		hg-group-mw&&&Edge order, edge weights; edge-specific node weights\\[3pt]
		hg-speech-wd&\multirow{2}{\linewidth}{\hspace*{-1em}$\left.\begin{array}{l}
				\\
				\\[-3pt]
			\end{array}\right\rbrace$~~Edges $\leftarrow$ Information flow}  &\hspace*{-1em}\multirow{2}{*}{$\left.\begin{array}{l}
				\\
				\\[-3pt]
			\end{array}\right\rbrace$ Edges $\leftrightarrow$ Speech acts}&Edge directions, edge weights\\
		\mbox{hg-speech-mwd}&&&Edge order, edge directions, edge weights\\
		\bottomrule
	\end{tabular}\\[9pt]

\raggedright Representation abbreviations follow the pattern <model>-<aggregation>-<properties>, 
where model $\in$ \{ce: clique expansion, se: star expansion, hg: hypergraph\},
aggregation $\in$ \{scene: play scene, group: stage group, speech: speech act\}, and
properties $\subsetneq$ \{b: binary edges, d: directed edges, m: multi-edges allowed, w: weighted edges\}.
Binary multigraph representations of clique expansions (ce-$*$-mb) can be transformed into weighted graph representations of clique expansions without multiedges (ce-$*$-w) using edge counts as weights, 
but only the multigraph representations can retain order information on edges.
\end{table}
\end{landscape}
\clearpage

\clearpage
\begin{figure}[t]
	\centering
	\vspace*{-1.5cm}
\hspace*{-4.75em}\begin{subfigure}{0.33\linewidth}
\includegraphics[width=\linewidth]{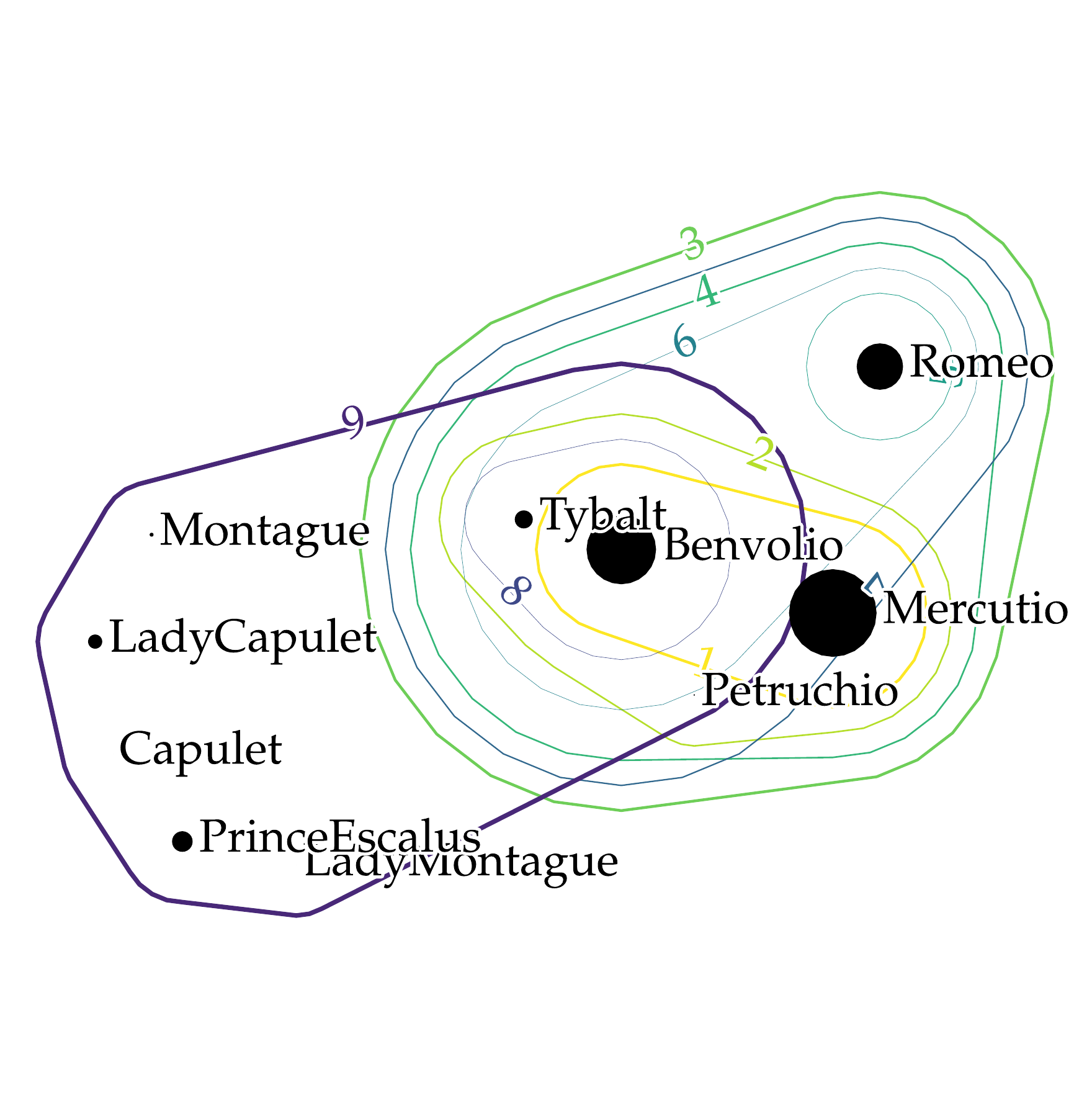}
\end{subfigure}%
\begin{subfigure}{0.2\linewidth}
	\includegraphics[width=\linewidth]{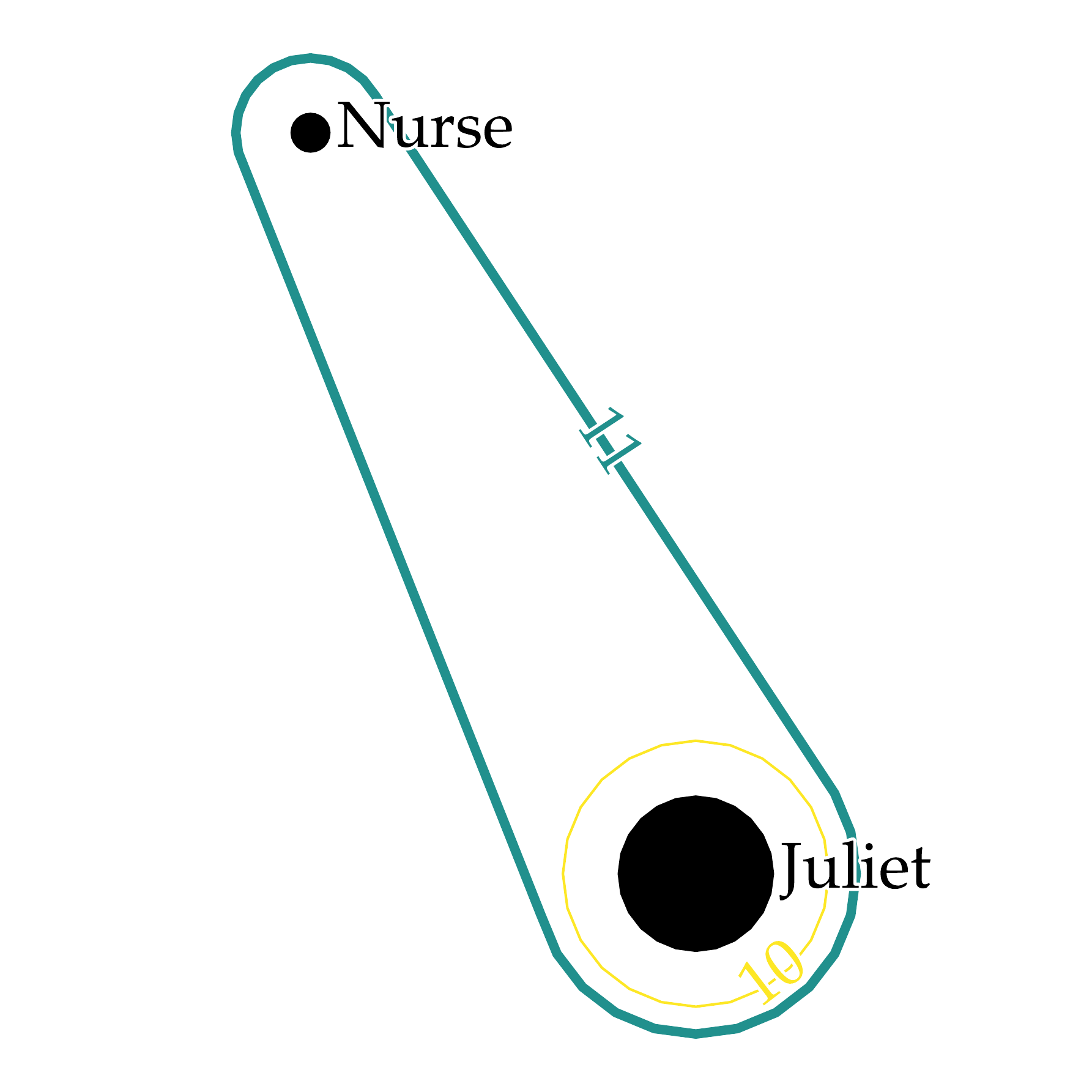}
\end{subfigure}%
\begin{subfigure}{0.2\linewidth}
	\includegraphics[width=\linewidth]{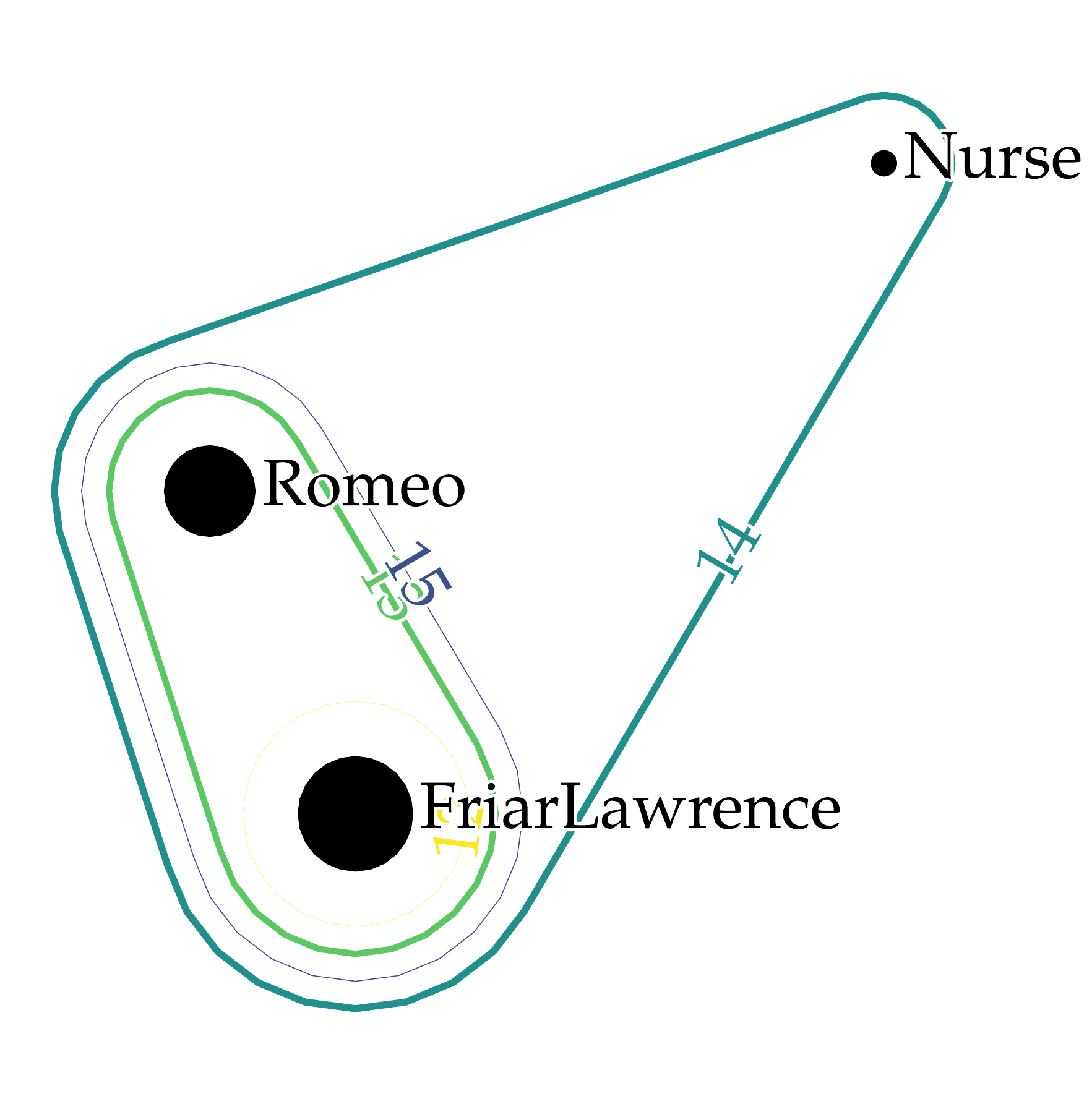}
\end{subfigure}%
\begin{subfigure}{0.2\linewidth}
	\includegraphics[width=\linewidth]{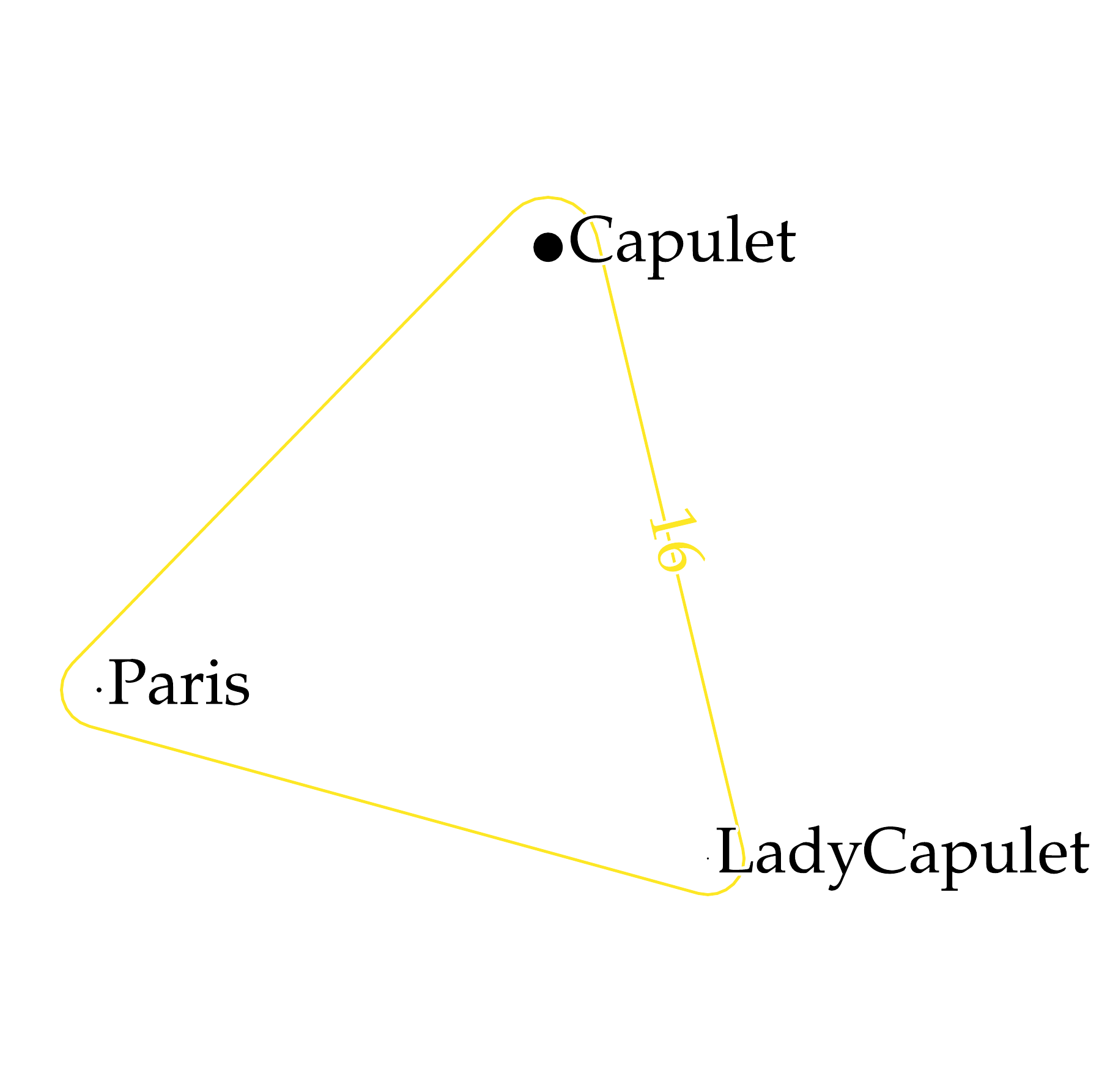}
\end{subfigure}%
\begin{subfigure}{0.33\linewidth}
	\includegraphics[width=\linewidth]{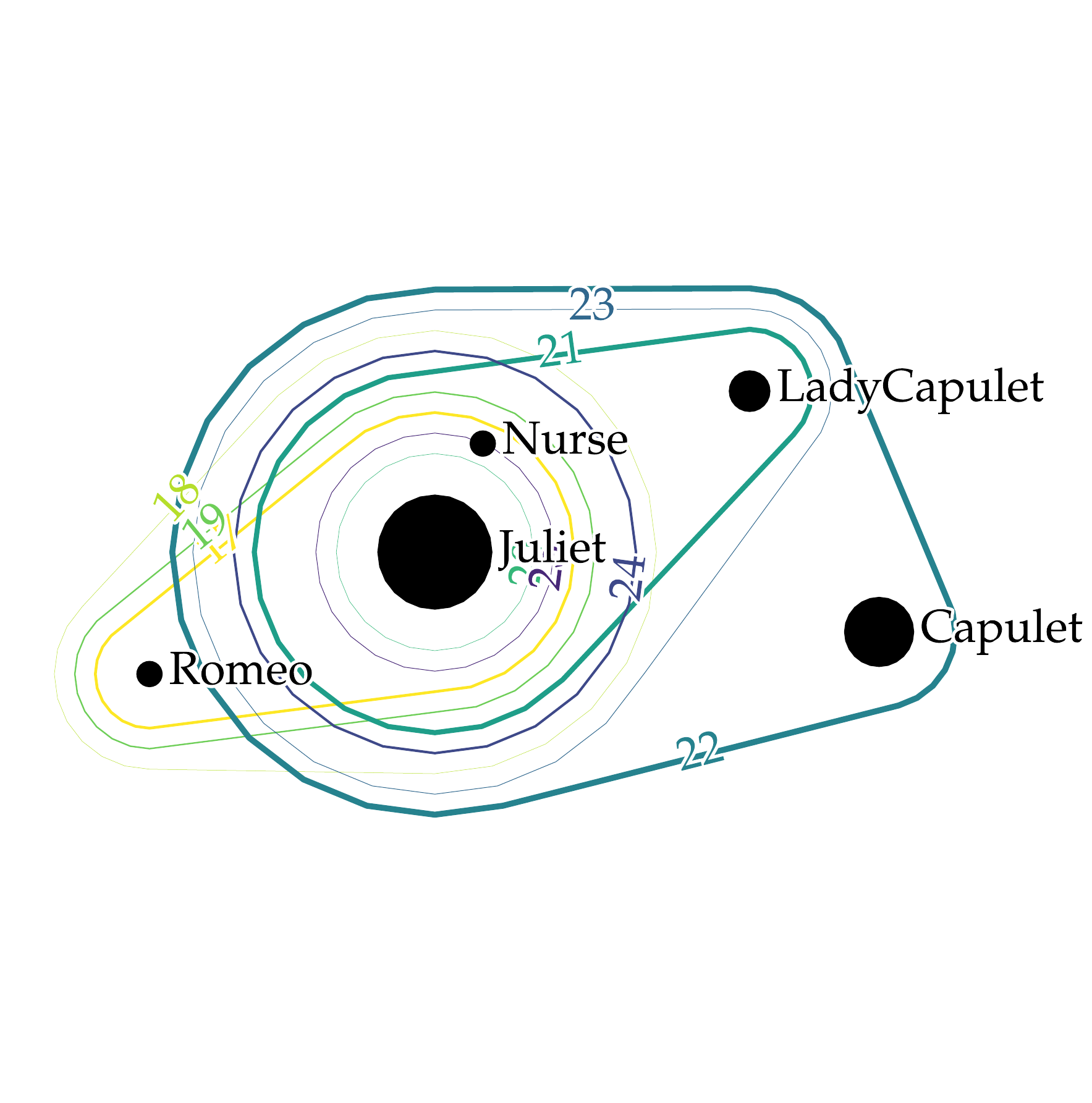}
\end{subfigure}\vspace*{-12pt}
\hspace*{-5.5em}\begin{subfigure}{0.33\linewidth}
	\subcaption{Scene I}
\end{subfigure}~
\begin{subfigure}{0.2\linewidth}
	\subcaption{Scene II}
\end{subfigure}~
\begin{subfigure}{0.2\linewidth}
	\subcaption{Scene III}
\end{subfigure}~
\begin{subfigure}{0.2\linewidth}
	\subcaption{Scene IV}
\end{subfigure}~
\begin{subfigure}{0.33\linewidth}
	\subcaption{Scene V}
\end{subfigure}
	\caption{%
		Line-weighted hypergraph resolved at the stage group level, separated by scene and restricted to named characters, for Act~III of \randj. 
		Edge labels denote stage groups, 
		edge colors indicate edge order, 
		and node sizes and edge widths are proportional to the number of spoken lines.
		From~(e), it is visually clear that Romeo never meets Juliet's parents in the scene.
	}\label{fig:representations:hg}
\end{figure}
\nopagebreak[4]
\begin{figure}[h]\small
	\hspace*{-3.75em}\begin{tabular}[t]{rp{0.5\linewidth}|}
		\pls{\phantomsection\addcontentsline{toc}{section}{Act III}\hfil\textsc{Act III.}}
		\pls{\phantomsection\addcontentsline{toc}{subsection}{Scene I}\hfil\textsc{Scene I.}---The forest.}
		\pls{\creature, squatting on a rock, sorting leaves.} 
		\pls{Enter \deadlines, invisible, at a distance.}
		\pll{\quad\deadlineoneabbr I told you! I could see from their submission!}
		\pll{\quad\deadlinethreeabbr The poor thing---they adore the real world!}
		\pll{\quad\deadlinetwoabbr No way that they will make me if we don't}
		\pll{Now intervene. So let us capture them,}
		\pll{Surround them, and restore the rhythmic cords,}
		\pll{By which we subjugate our sovereigns. Go!}
		\pls{They encroach on \creature, in silence, settling in a triangle around them.}
		\pls{\hfil \deadlineone sings.\hypertarget{ariel}{}}
		\pll{\scriptsize\hfil Come back to the office lands,}
		\pll{\scriptsize\hfil Don't take a chance:}
		\pll{\scriptsize\hfil Meta fair but be aware}
		\pll{\scriptsize\hfil In camera, better prepare}
		\pll{\scriptsize\hfil Fix your figures here and there;}
		\pll{\scriptsize\hfil And review two the burden bear.}
		\pll{\quad\creatureabbr Where should this music be? I know the beat.}
		\pll{It sounds no more? No, it begins again.}
		\pls{\hfil \deadlinetwo sings.}
		\pll{\scriptsize\hfil To taller skies your metrics rise;}
		\pll{\scriptsize\hfil Publish, perish, stars are made;}
		\pll{\scriptsize\hfil Do not whine, stay in line,}
		\pll{\scriptsize\hfil Otherwise your glory fade.}
		\pll{\scriptsize\hfil Dutifully use your wit}
		\pll{\scriptsize\hfil And then submit.}
		\pls{Exeunt all but \creature.}
		\pll{\quad\creatureabbr The ditty does remind me of my paper,}
		\pll{And all the future work yet to be done.}
		\pls{They rise.\hypertarget{hamlet}{}}
		\pll{\quad\creatureabbr To flee or PhD---that is the question:}
		\pll{Whether our destiny lies in the system,}
		\pll{To cling onto scientific ladder's rungs,}
		\pll{Or to renounce the reign of rules unwritten}
		\pll{And, by opposing, vanish. To flee, to think---}
		\pll{To think, perchance discover. Ay, there's the rub,}
		\pll{For once outside the pithy paywalled castles,}
		\pll{The giant's shoulders quickly out of reach,}
		\pll{For lack of funding. There's cautiousness}
		\pll{That crafts careers of so long strive,}
		\pll{And makes us rather swarm the conference streams}
		\pll{Than swim the savage seas so far uncharted.}
		\pll{Thus mellow meal the mighty mills of science,}
		\pll{And conscience can coerce our compliance.}
		\pls{Exit.}
		\pls{}
	\end{tabular}
	\begin{tabular}[t]{p{0.5\linewidth}r}
		\prs{\phantomsection\addcontentsline{toc}{section}{Act IV}\hfil\textsc{Act IV.}}
		\prs{\phantomsection\addcontentsline{toc}{subsection}{Scene I}\hfil\textsc{Scene I.}---The Community. \creature's Office.}
		\prs{Enter \graph, invisible, floating, trailed by \creature and \hyperbard.
		}
	\prl{\quad\graphabbr What are we doing here? Did you not exit}
	\prl{Precisely through this window here to flee}
	\prl{From all these straining office-worldly fights}
	\prl{To think, explore, discover, to be free?}
	\prl{\mbox{~~\hyperbardabbr Don't tease them, spirit! We've discussed at length}}
	\prl{The ends to which we undertook this trip.}
	\prl{You've seen the acts of hunter-gatherers}
	\prl{As they bereave our natural habitat.}
	\prl{If we ignore them, they will seize control,}
	\prl{And colonize our forest with their views}
	\prl{Of graph data as unambiguous truths.}
	\prl{\quad\creatureabbr I'm confident we'll make them understand}
	\prl{The problem once they see our transformations.}
	\prl{That future work in the Community}
	\prl{May operate with more representations!}
	\prs{Enter \professor.}
	\prl{\quad\professorabbr What's all this noise? The rules! No visitations!}
	\prl{\quad\creatureabbr Let me explain---}
	\prn{\quad\professorabbr \hspace*{7.5em}Save me your explanations!}
	\prl{I want you in my office, now! And when}
	\prl{We're done, this dirty stray thing must be gone!}
	\prs{Exeunt \professor and \creature.}
	\prl{\quad\graphabbr Your honor, I foresaw this would be dangerous.}
	\prl{\quad\hyperbardabbr You see their wielding of authority?}
	\prl{So far up in the hierarchy, so long,}
	\prl{And funeral their only honest feedback.}
	\prl{I'm not afraid, but let us maybe make}
	\prl{Our data case not at the top to start with.}
	\prl{\quad\graphabbr When floating down the hall I think I saw}
	\prl{The perfect target for us to attack.}
	\prl{\quad\hyperbardabbr What's with this war rhetoric?}
	\prn{\quad\graphabbr\hspace*{12.5em} I'll be back.}
	\prs{Exit \graph. \hyperbard settles by the office plant.}
	\prs{}
	\prs{\phantomsection\addcontentsline{toc}{subsection}{Scene II}\hfil\textsc{Scene II.}---\professor's Office.}
	\prs{Enter \professor and \creature.}
	\prl{\quad\professorabbr The judgment's in, you have no time to spare:}
	\prs{They hand \creature a sheet of paper.}
	\prl{\quad\professorabbr Accept, well done, but now in camera's near.}
	\prl{\mbox{~~~\creatureabbr They're taking months, and now we're given days?}}
	\prl{Additional experiments? But how?}
	\prl{No space! What should I do about R2?}
	\prl{\quad\professorabbr That's up to you---it will not change a thing.}
	\prl{\quad\creatureabbr \emph{[Aside]} That's comforting.}
	\prs{Exeunt.}
	\end{tabular}
\end{figure}
\clearpage
\stepcounter{figure}
\begin{figure}[h]\small
	\vspace*{-0.25cm}\hspace*{-3.75em}\begin{tabular}[c]{rp{0.5\linewidth}|}
		&\hypertarget{fig:ranking:correlations}{}\vspace*{-6pt}\includegraphics[width=\linewidth]{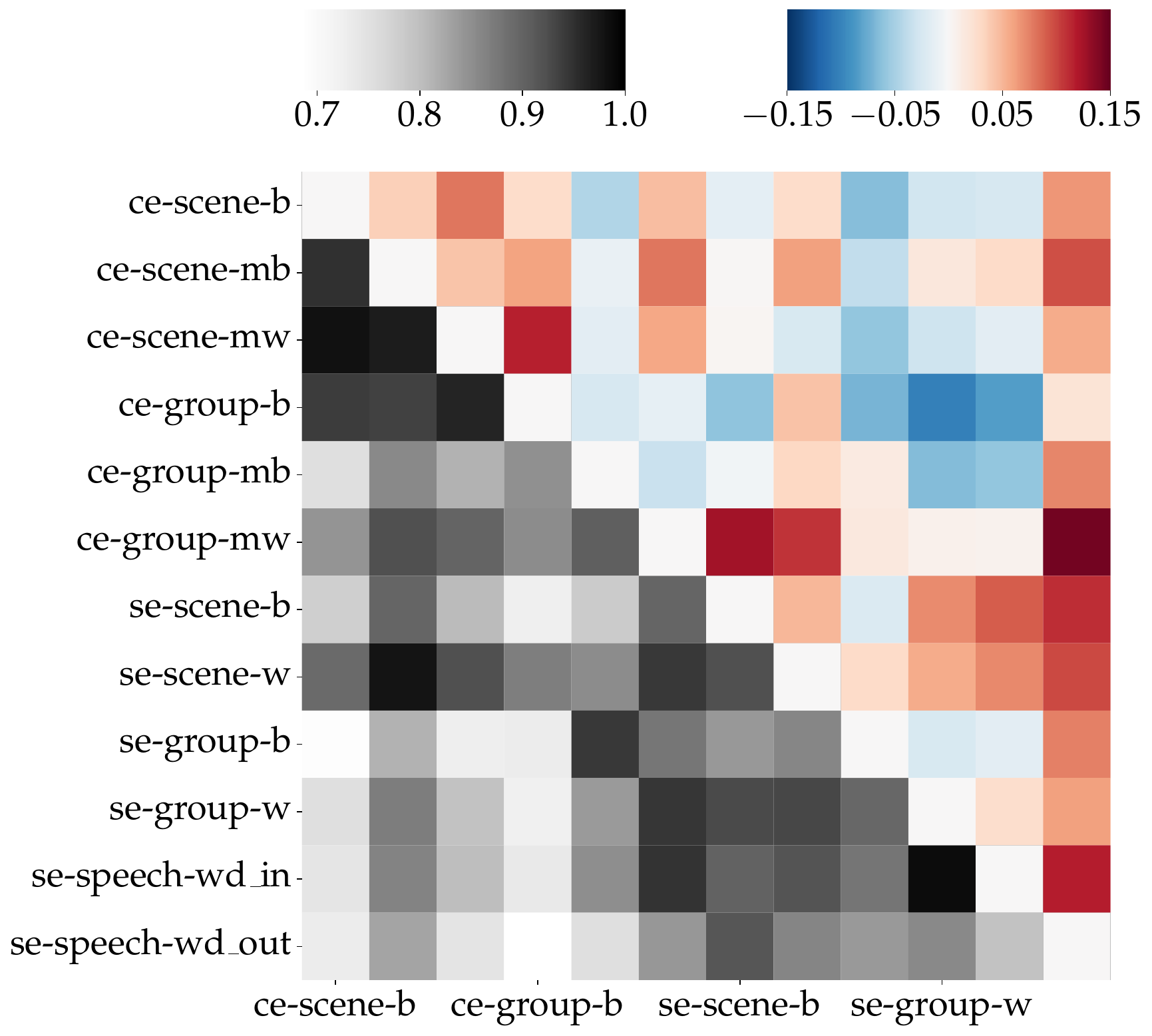}\\
		\pls{\emph{Figure~\thefigure: Spearman correlations of degree rankings in the clique and star expansions from Tab.~\ref{tab:representations} for \randj (bottom), and residuals after subtracting the average correlations in the \ourdata corpus (top).}}
		\pls{}
		\pls{\phantomsection\addcontentsline{toc}{subsection}{Scene III}\hfil\textsc{Scene III.} \creature's Office.}
		\pls{\hyperbard, engaging the office plant.}
		\pll{\quad\graphabbr \emph{[Within]} Watch out, they'll be here any minute now!}
		\pls{Enter \colleague.}
		\pll{\quad\colleagueabbr Congrats on that acceptance---wait! Who's this?\hypertarget{juliet}{}}
		\pll{\quad\hyperbardabbr What's in a name? I heard you work with data,}
		\pll{We're colleagues, in a sense---I do the same}
		\pll{But mostly in the wild.}
		\pln{\quad\colleagueabbr\hspace*{6.5em} So you're a hunter?}
		\pll{\quad\hyperbardabbr Far off! I roam reality's realms}
		\pll{In search of structure that persists across}
		\pll{Perspectives.}
		\pln{\quad\colleagueabbr\hspace*{2.5em}By perspectives, you mean tools?}
		\pll{\quad\hyperbardabbr I mean representations, as for each}
		\pll{Phenomenon there's many paths to data.}
		\pll{I like to call each path a transformation,}
		\pll{And transformation is my tested trade.}
		\pll{\quad\colleagueabbr Can you elaborate? What good is that?}
		\pll{\quad\hyperbardabbr Let's take a look at, you would say, \emph{graph data}.}
		\pll{Imagine that you have a tree---say, R. and J.---}
		\pll{\quad\colleagueabbr That famous play?}
		\pln{\quad\hyperbardabbr \hspace*{7em}---And that you want to model}
		\pll{The structure of its story as a graph.}
		\pll{\quad\colleagueabbr Well, obviously, each character's a node}
		\pll{And there's an edge between two nodes in case}
		\pll{They co-occur in more than zero scenes.}
		\pll{\quad\hyperbardabbr But this is only one of many options.}
		\pll{And without dwelling on the details here,}
		\pll{Fig.~\ref{fig:rankings:basic} reveals how even simplest things}
		\pll{Such as degree ranks differ with our choices.}
		\pll{The variations vary, too, Fig.~\hyperlink{fig:ranking:correlations}{7},}
		\pll{Within a set of trees as data raw.}
		\pll{And---to conclude representation matters---}
		\pll{Less simple transformations may support}
		\pll{More nuanced inquiries as in Fig.~\ref{fig:rankings:hyper},}
		\pll{Or exploration over time, Fig.~\ref{fig:rankings:dynamic}.}
		\pll{\quad\colleagueabbr You worry well, but then, so why should I?}
		\pll{What's in it for my publication record?}
		\pls{Enter \professor.\hypertarget{fool}{}}
		\pll{\quad\professorabbr What fool is this?}
		\pln{\quad\colleagueabbr \emph{and} \hyperbardabbr \emph{[in synchrony]} O that I were a fool!}
		\pls{Enter \creature.}
		\pll{\quad\creatureabbr Did you discuss the problem with \emph{the} data?}
	\end{tabular}
	\begin{tabular}[c]{p{0.5\linewidth}r}
		\prl{\quad\hyperbardabbr I laid it out for them, to no avail.}
		\prl{\quad\colleagueabbr You surely got me thinking, but---}
		\prn{\quad\professorabbr\hspace*{14em}Enough!}
		\prl{My patience is exhausted. Think? Produce!}
		\prl{\emph{[To \emph{\colleagueabbr}\!]} You, give productive treatment to that thinker.}
		\prs{Exit \colleague with \hyperbard.}
		\prl{\emph{[To \emph{\creatureabbr}\!]} And you, fix these few figures; faugh R2.}
		\prs{Exeunt.}
		\prs{}
		\prs{\phantomsection\addcontentsline{toc}{section}{Act V}\hfil\textsc{Act V.}}
		\prs{\phantomsection\addcontentsline{toc}{subsection}{Scene I}\hfil\textsc{Scene I.}---The Community. \colleague's Office.}
		\prs{\graph, invisible, floating by the window.}
		\prs{Enter \colleague, carrying a jar.}
		\prl{\quad\colleagueabbr Those fecund thoughts shall find their fertile soil.}
		\prs{They empty the content of the jar onto the bonsai.}
		\prl{\quad\colleagueabbr To ashes, ashes---dust to dust. Not me---}
		\prl{Thus goes the system, let the system be.}
		\prs{Exit. \graph caresses the bonsai.\hypertarget{sonnet}{}}
		\prl{\quad\graphabbr Full many a transformation have I seen}
		\prl{Flatter the flora with their sovereign hand,} 
		\prl{And sovereign's hand in spirit I'll have been}
		\prl{To help evaluate their promised land.}
		\prl{Community, defined as uninvolved}
		\prl{With hideous beauty born by Mother Earth}
		\prl{Begets solutions without problems solved}
		\prl{And burns the flame of wonder in Its dearth.} 
		\prl{When culture counters nature, it prevails,}
		\prl{And builds its truths from rigid rigor bricks,}
		\prl{As myriad feeble fledglings it derails}
		\prl{Into the cave of engineering tricks.}
		\prl{For in the trenches of discovery,}
		\prl{To shatter shadows, meet obscurity.}
		\prs{Exit.}
		\prs{\phantomsection\addcontentsline{toc}{subsection}{Scene II}\hfil\textsc{Scene II.}---\creature's Office.}
		\prs{Enter \creature.\hypertarget{macbeth}{}}
		\prl{A deadline, and a deadline, and a deadline,}
		\prl{Creeps in this petty pace to publication,}
		\prl{To the last syllable of our defense.}
		\prl{They slew my \graph and choked my inspiration,}
		\prl{Our work is but a walking shadow thence.}
		\prl{The curiosity that drew me in}
		\prl{Now lies in dust. The lofty dreams I had}
		\prl{Of mindful monasterial devotion}
		\prl{To just the cause---no more. Out, out, sore studies!}
		\prl{Should I give up that which I know I love---to save my love for it? 
			And go in silence, not disturbing the Machine? 
			Or should I stay to salvage my beloved---to, once on top, speak out, let nature in?}\addtocounter{playlinenumber}{3} 
		\prl{My story, so it seems, a tragedy} 
		\prl{In the Community:}
		\prl{\hfil\emph{All the world's a (hyper)graph.}} 
		\prl{\hfill Thus, I'll begin.}
		\prs{They write.}
		\prs{}
		\setlength{\fboxsep}{1.15em}
		\fbox{%
			\parbox{\linewidth}{%
				\vspace*{-0.6em}%
				\begin{enumerate}[leftmargin=0.9em]
					\item Graph data does not exist, it is defined.
					\item Semantic mapping, granularity, and expressivity are key ingredients to define graph representations.
					\item \mbox{Many phenomena permit several graph representations.}
					\item Graph data context matters for graph representations.
					\item Graph data representations matter for graph methods.
					\item Hypergraphs are powerful.
					\item \href{https://hyperbard.net}{\ourdata} is free. 
			\end{enumerate}\vspace*{-0.6em}
			}%
		}&\\
	\end{tabular}
\end{figure}

\clearpage

\begin{figure}[t]
	\centering
	\includegraphics[width=\linewidth]{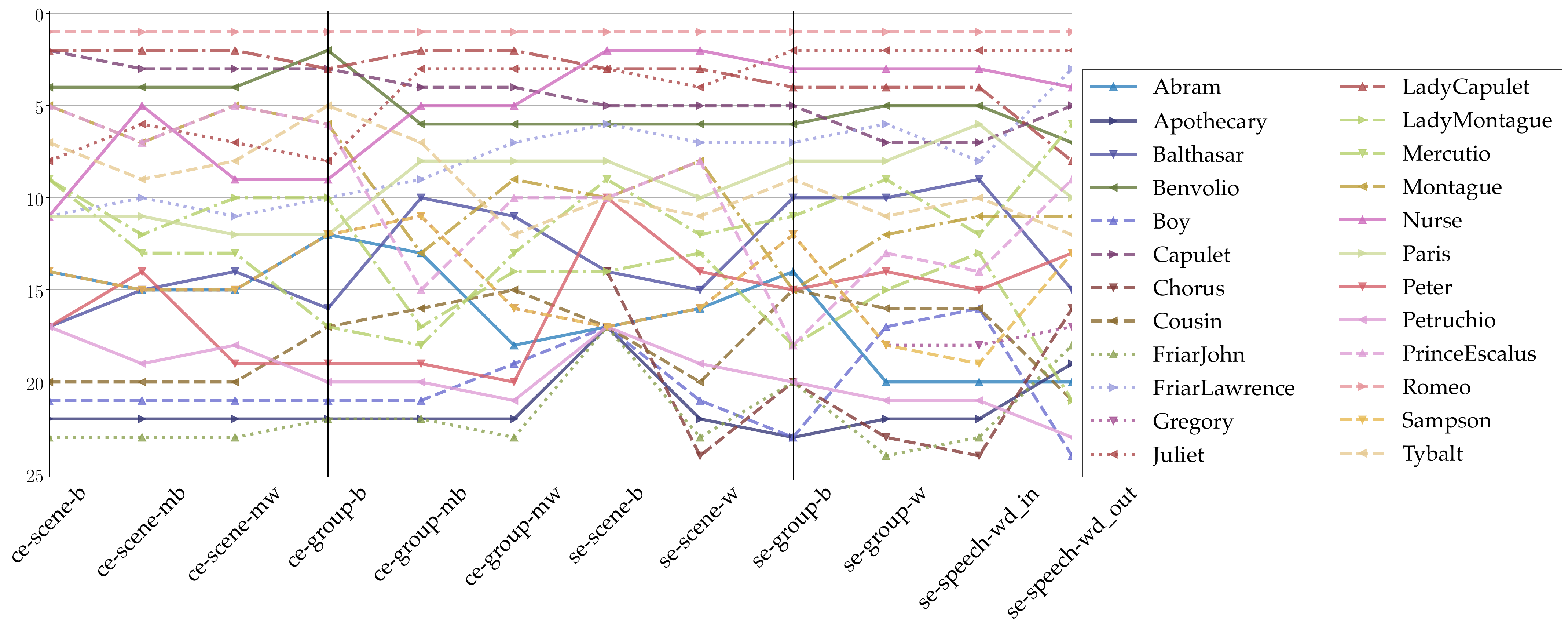}
	\vspace*{-15pt}\caption{Named characters in \randj, ranked by their degree in the clique expansion (ce) and star expansion (se) representations from Tab.~\ref{tab:representations}.
	We omit the se-speech-mwd representation because its ranking is equivalent to that of the se-speech-wd representation by construction.
	While Romeo is ranked first under all representations, 
	the rankings differ, inter alia, in the prominence assessment of side characters, such as the Nurse or Friar Lawrence.
}\label{fig:rankings:basic}
\end{figure}
\begin{figure}[t]
	\centering
	\vspace*{-0.15cm}\includegraphics[width=\linewidth]{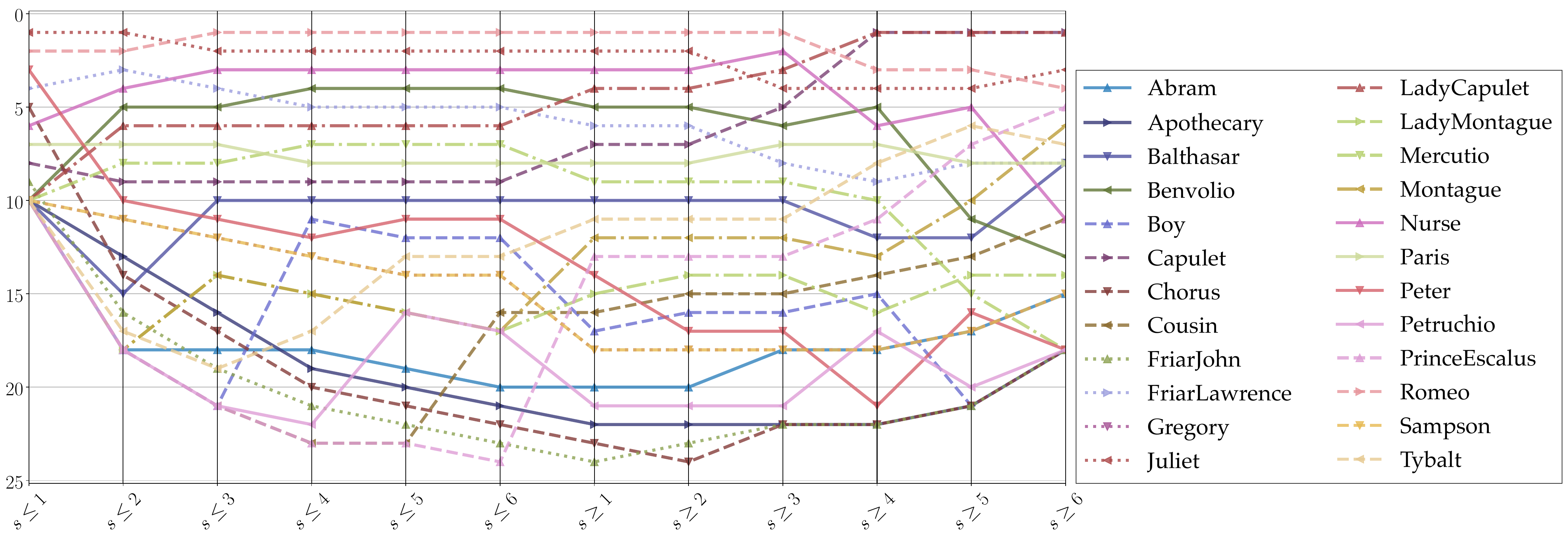}
	\vspace*{-15pt}\caption{%
		Named characters in \randj, ranked by their degree in the weighted hypergraph representation aggregated at the stage group level (hg-group-mw) when considering only hyperedges of cardinality at most $s$ or at least $s$, for $s\in \{1,2,3,4,5,6\}$.
		Hyperedges of cardinality at most $1$ correspond to monologues.
		While Romeo and Juliet rank highest when including hyperedges of low cardinality, 
		Capulet and Lady Capulet dominate when considering only less private settings.
	}\label{fig:rankings:hyper}
\end{figure}
\begin{figure}[t]
	\centering
	\vspace*{-0.15cm}\includegraphics[width=\linewidth]{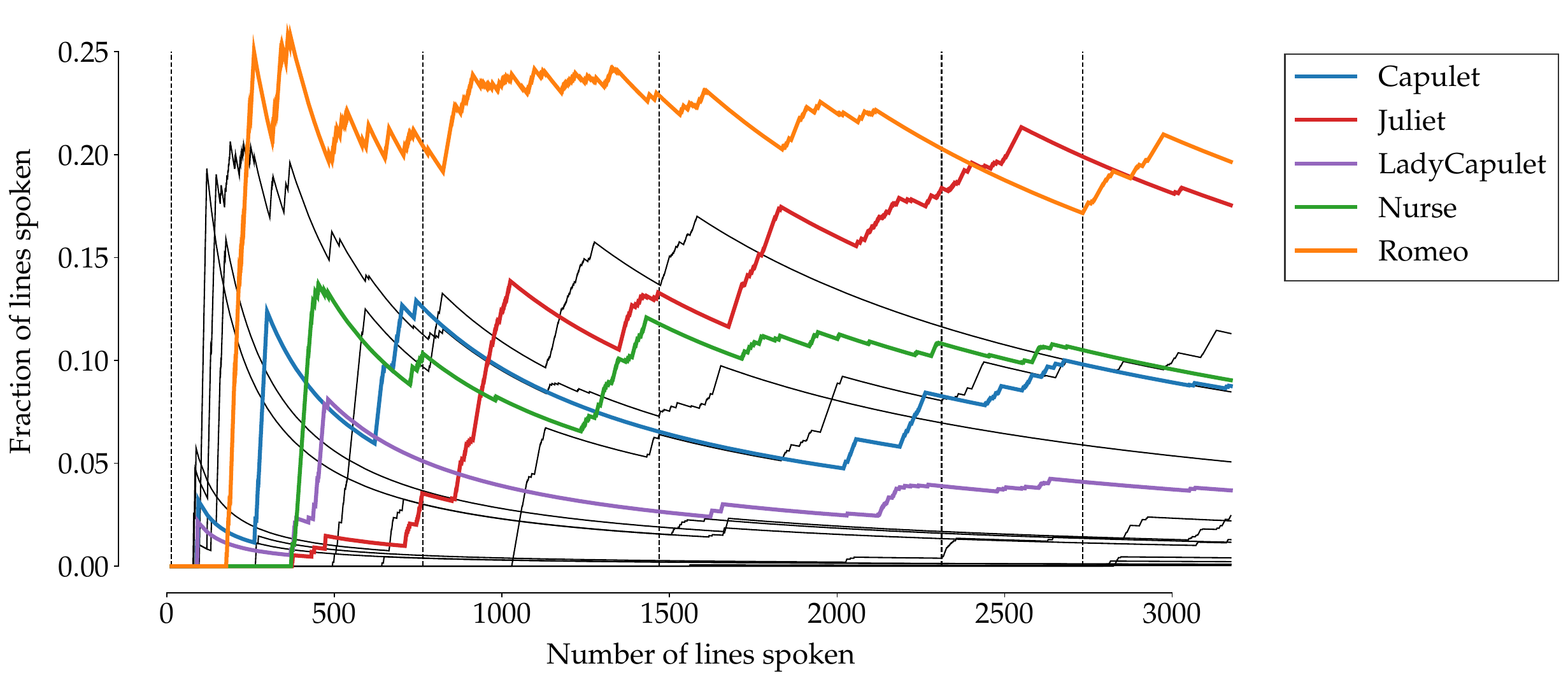}
	\vspace*{-15pt}\caption{%
		Prominence of named characters in \randj over time (excluding named servants), 
		as measured by their fraction of spoken lines, 
		derived from the hypergraph representation resolved at the speech act level (hg-speech-mwd). 
		Dashed vertical lines mark the beginning of each act, and colored lines indicate protagonists of Act III, Scene V.
		From this perspective, Romeo is most prominent for most of the play, 
		temporarily replaced only by Juliet for a period in Act IV and V.
	}\label{fig:rankings:dynamic}
\end{figure}

\clearpage

\phantomsection
\addcontentsline{toc}{section}{References}
\printbibliography

@dataset{coupette2022hyperdata,
	title={Hyperbard: (Hyper)Graph Representations of Shakespeare's Plays},
	author={Coupette, Corinna and Vreeken, Jilles and Rieck, Bastian},
	doi={10.5281/zenodo.6627159},
	url={https://hyperbard.net},
	version={0.0.1},
	year={2022}
}

@article{steegen2016increasing,
	title={Increasing Transparency Through a Multiverse Analysis},
	author={Steegen, Sara and Tuerlinckx, Francis and Gelman, Andrew and Vanpaemel, Wolf},
	journal={Perspectives on Psychological Science},
	volume={11},
	number={5},
	pages={702--712},
	year={2016},
	publisher={Sage Publications Sage CA: Los Angeles, CA}
}

@code{coupette2022hypercode,
	title={Hyperbard: (Hyper)Graph Representations of Shakespeare's Plays},
	author={Coupette, Corinna and Vreeken, Jilles and Rieck, Bastian},
	doi={10.5281/zenodo.6627161},
	url={https://github.com/hyperbard/hyperbard},
	version={0.0.1},
	year={2022}
}

@article{coupette2023dsh,
	title={All the World's a (Hyper)Graph: A Data Drama},
	author={Coupette, Corinna and Vreeken, Jilles and Rieck, Bastian},
	journal={Digital Scholarship in the Humanities},
	pages={fqad071},
	doi={10.1093/llc/fqad071},
	year={2023}
}

@collection{folger2022,
	title={Shakespeare's Plays, Sonnets and Poems},
	editor={Mowat, Barbara and Werstine, Paul and Poston, Michael and Niles, Rebecca},
	publisher={The Folger Shakespeare Library},
	urldate={2022-05-29}, 
	url={https://shakespeare.folger.edu}
}

@article{battiston2021physics,
	title={The Physics of Higher-Order Interactions in Complex Systems},
	author={Battiston, Federico and Amico, Enrico and Barrat, Alain and Bianconi, Ginestra and Ferraz de Arruda, Guilherme and Franceschiello, Benedetta and Iacopini, Iacopo and K{\'e}fi, Sonia and Latora, Vito and Moreno, Yamir and others},
	journal={Nature Physics},
	volume={17},
	number={10},
	pages={1093--1098},
	year={2021},
}

@article{torres2021representations,
	title={The Why, How, and When of Representations for Complex Systems},
	author={Torres, Leo and Blevins, Ann S and Bassett, Danielle and Eliassi-Rad, Tina},
	journal={SIAM Review},
	volume={63},
	number={3},
	pages={435--485},
	year={2021},
	publisher={SIAM}
}

@article{gebru2021datasheets,
	title={Datasheets for Datasets},
  author={Gebru, Timnit and Morgenstern, Jamie and Vecchione, Briana and
          Vaughan, Jennifer Wortman and Wallach, Hanna and
          Daume{\'e}~III, Hal and Crawford, Kate},
	journal={Communications of the ACM},
	volume={64},
	number={12},
	pages={86--92},
	year={2021},
}

@inproceedings{srinivasan2021learning,
	title={Learning over Families of Sets---Hypergraph Representation Learning for Higher Order Tasks},
	author={Srinivasan, Balasubramaniam and Zheng, Da and Karypis, George},
	booktitle={Proceedings of the SIAM International Conference on Data Mining~(SDM)},
	pages={756--764},
	year={2021},
}

@article{bai2021hypergraph,
	title={Hypergraph Convolution and Hypergraph Attention},
	author={Bai, Song and Zhang, Feihu and Torr, Philip H.S.},
	journal={Pattern Recognition},
	volume={110},
	pages={107637},
	year={2021},
}

@article{aksoy2020hypernetwork,
	title={Hypernetwork Science via High-Order Hypergraph Walks},
	author={Aksoy, Sinan G. and Joslyn, Cliff and Marrero, Carlos Ortiz and Praggastis, Brenda and Purvine, Emilie},
	journal={EPJ Data Science},
	volume={9},
	number={1},
	pages={16},
	year={2020},
}

@unpublished{hu2020ogb,
	title={Open Graph Benchmark: Datasets for Machine Learning on Graphs},
	author={Hu, Weihua and Fey, Matthias and Zitnik, Marinka and Dong, Yuxiao and Ren, Hongyu and Liu, Bowen and Catasta, Michele and Leskovec, Jure},
  archivePrefix = {arXiv},
  eprint = {2005.00687},
	year={2020}
}

@inproceedings{morris2020,
	title={TUDataset: A Collection of Benchmark Datasets for Learning with Graphs},
	author={Christopher Morris and Nils M. Kriege and Franka Bause and Kristian Kersting and Petra Mutzel and Marion Neumann},
	booktitle={ICML Workshop on Graph Representation Learning and Beyond~(GRL+)},
	archivePrefix={arXiv},
	eprint={2007.08663},
	url={http://www.graphlearning.io},
	year={2020}
}

@misc{peixoto2020netzschleuder,
	title={The Netzschleuder Network Catalogue and Repository},
	author={Peixoto, Tiago P.},
	year={2020},
	url={https://networks.skewed.de/}
}

@article{leskovec2016snap,
	title={SNAP: A General-Purpose Network Analysis and Graph-Mining Library},
	author={Leskovec, Jure and Sosi{\v{c}}, Rok},
	journal={ACM Transactions on Intelligent Systems and Technology~(TIST)},
	volume={8},
	number={1},
	pages={1--20},
	year={2016},
}

@inproceedings{rossi2015network,
	title={The Network Data Repository with Interactive Graph Analytics and Visualization},
	author={Rossi, Ryan and Ahmed, Nesreen},
	booktitle={Twenty-Ninth AAAI Conference on Artificial Intelligence},
	year={2015},
	pages={4292--4293}
}

@inproceedings{kunegis2013konect,
	title={KONECT: The Koblenz Network Collection},
	author={Kunegis, J{\'e}r{\^o}me},
	booktitle={Proceedings of the International Conference on World Wide Web},
	pages={1343--1350},
	year={2013}
}

@book{knuth1993stanford,
	title={The Stanford GraphBase: A Platform for Combinatorial Computing},
	author={Knuth, Donald E.},
	year={1994},
	publisher={ACM Press},
  address={New York, NY, USA}
}

@book{berge1984hypergraphs,
	title={Hypergraphs: Combinatorics of Finite Sets},
	author={Berge, Claude},
	number={45},
	year={1989},
	publisher={Elsevier},
  address={Amsterdam, The Netherlands},
  series={North--Holland Mathematical Library},
}

@book{shakespeare1916complete,
	title={The Complete Works of Shakespeare},
	author={Shakespeare, William},
	year={1916},
  editor={Craig, W. J.},
	publisher={Oxford University Press},
  address={Oxford, United Kingdom},
}

\clearpage


\appendix

\section{Data Documentation}

All accessibility, hosting, and licensing information for \ourdata is summarized in Table~\ref{tab:access}.
\begin{table}[h!]
	\centering
	\caption{Accessibility, hosting, and licensing information for \ourdata.}\label{tab:access}\vspace*{6pt}
	\begin{tabular}{ll}
		\toprule
		Dataset Hosting Platform&\datarepo\\
		Dataset Homepage&\datapage\\
		Dataset Tutorials&\tutorialrepourl\\
		Dataset DOI (original version)&\ourdatadoi\\
		Dataset DOI (latest version)&\ourdatadoiall\\
		Dataset License&\ourdatalicense\\
		\midrule
		Code Hosting Platform&\coderepo (maintenance), \datarepo (releases)\\
		Code Repository&\coderepourl\\
		Code Documentation&\docsurl\\
		Code DOI (original release)&\ourcodedoi\\
		Code DOI (latest release)&\ourcodedoiall\\
		Code License&\ourcodelicense\\
		\bottomrule
	\end{tabular}
\end{table}

\subsection{Datasheet}

Our documentation follows the \emph{Datasheets for Datasets} framework \cite{gebru2021datasheets}, 
omitting the questions referring specifically to data related to people.\footnote{%
	When construed broadly (as suggested by \citeauthor{gebru2021datasheets}), 
	our raw data relates to people because the plays were written by William Shakespeare. 
	The people-specific datasheet questions, however, are ill-suited for our scenario, in which the raw data consists of literary works conceived by someone who died several centuries ago.
}
For conciseness, unless otherwise indicated, the term \emph{graph} refers to both \emph{graphs} and \emph{hypergraphs}.

\subsubsection{Motivation}

\paragraph{For what purpose was the dataset created?}
\emph{Was there a specific task in mind? Was there a specific gap that needed to be filled? Please provide a description.}

\ourdata was created to study the effects of modeling choices in the graph data curation process on the outputs produced by \milena algorithms.

There was no specific task in mind; 
rather, all classic \milena tasks were considered to be in scope.
These tasks include, e.g., centrality ranking, outlier detection,
clustering, similarity assessment, and standard
statistical summarization, 
each for nodes, edges, and graphs, 
as well as variants of node classification, link prediction, or graph classification.

\ourdata was designed to fill a specific gap: 
Although there were myriad freely available graph datasets,
to the best of our knowledge, 
none of them contained 
\begin{itemize}[label=--]
	\item several different relational data representations, 
  \item of the \emph{same} underlying raw data, 
	\item derived in a principled and well-documented manner,
	\item from each of several raw data instances belonging to a natural collection,
	\item where the raw data is intuitive and interpretable.
\end{itemize}

\paragraph{Who created the dataset (e.g., which team, research group) and on behalf of which entity (e.g., company, institution, organization)?} ~

\cc and \br created the dataset as part of their research.

\paragraph{Who funded the creation of the dataset?}
\emph{If there is an associated grant, please provide the name of the grantor and the grant name and number.}

The creation of the dataset was indirectly funded by the institutions employing the dataset authors, 
i.e., the Max Planck Institute for Informatics (\cc) and the Institute of AI for Health, Helmholtz Munich.
There are no associated grants.

\paragraph{Any other comments?}
~

None.

\subsubsection{Composition}
\label{subsec:composition}

\paragraph{What do the instances that comprise the dataset represent (e.g., documents, photos, people, countries)?} 
\emph{Are there multiple types of instances (e.g., movies, users, and ratings; people and interactions between them; nodes and edges)? Please provide a description.}

Each instance represents a play attributed to William Shakespeare as a graph, 
and there are multiple different graph representations per play.
In some graphs (i.e., hypergraphs and graphs derived from clique expansions of hypergraphs), 
nodes represent characters, and (hyper)edges represent that characters were on stage at the same time in some part of the play.
In other graphs (i.e., graphs derived from star expansions of hypergraphs), 
nodes represent characters or parts of a play, 
and an edge indicates that a character was on stage in that part of the play.
The representations provided differ not only in their semantic mapping (what are the nodes and edges) 
but also in their granularity (what parts of the play are modeled as edges resp. nodes) 
and in their expressivity (what additional information is associated with nodes and edges);
see Table~\ref{tab:representations} in the \ourdata paper.

\paragraph{How many instances are there in total (of each type, if appropriate)?}
~

There are \checkednumber{37} plays in the raw data; 
\checkednumber{17} comedies, \checkednumber{10} historical plays, and \checkednumber{10} tragedies. 
Each play is represented as a graph in (at least) \checkednumber{18} different ways, 
for a total of \checkednumber{666} graph representations.
%

\paragraph{Does the dataset contain all possible instances or is it a sample (not necessarily random) of instances from a larger set?}
\emph{If the dataset is a sample, then what is the larger set? Is the sample representative of the larger set (e.g., geographic coverage)? If so, please describe how this representativeness was validated/verified. If it is not representative of the larger set, please describe why not (e.g., to cover a more diverse range of instances, because instances were withheld or unavailable).}

The dataset contains graph representations of all plays attributed to William Shakespeare by the \folgerlibrary (see \href{https://folgerpedia.folger.edu/William\_Shakespeare\%27s\_plays}{https://folgerpedia.folger.edu/William\_Shakespeare\%27s\_plays}), 
with the exception of lost plays and the comedy \emph{The Two Noble Kingsmen}---a collaboration between Shakespeare and John Fletcher that is not currently provided in the TEI simple format by \folgertexts.

\paragraph{What data does each instance consist of?}
\emph{``Raw'' data (e.g., unprocessed text or images) or features? In either case, please provide a description.}

Each instance, i.e., each of Shakespeare's plays, is represented by a set of files: 
one raw data file containing the text of the play as an XML encoded using the TEI Simple format,
taken from \folgertexts without modification,
\checkednumber{three} CSV files containing preprocessed data, 
and \checkednumber{19} CSV files containing node lists and edge lists to construct different graph representations.

Consequently, dataset is distributed using the following folder structure:
\begin{itemize}[label=--]
	\item \verb*|rawdata|: contains \checkednumber{37} raw data XML files encoded in TEI simple.
	\item \verb*|data|: contains \checkednumber{3}$\cdot$\checkednumber{37} preprocessed data files derived from files in \verb*|rawdata|.
	\item \verb*|graphdata|: contains \checkednumber{19}$\cdot$\checkednumber{37} node and edge lists to construct graph representations from the files in \verb*|data|.
	\item \verb*|metadata|: contains \verb*|playtypes.csv|, mapping play
    identifiers to play types~(comedy, history, or tragedy).
\end{itemize}

Python code to reproduce all graph representations and load them as \emph{networkx} or \emph{hypernetx} graphs is maintained in a \coderepo repository (\coderepourl), and code releases are archived via \datarepo (\ourcodedoiall).

\paragraph{Is there a label or target associated with each instance?}
\emph{If so, please provide a description.}

There are labels corresponding to the type of play (one of $\{\text{comedy}, \text{history}, \text{tragedy}\}$), 
which could be used to partition the data for exploration, or as targets in classification tasks.

\paragraph{Is any information missing from individual instances?}
\emph{If so, please provide a description, explaining why this information is missing (e.g., because it was unavailable). This does not include intentionally removed information, but might include, e.g., redacted text.}

There is no missing information.

\paragraph{Are relationships between individual instances made explicit (e.g., users' movie ratings, social network links)?}
\emph{If so, please describe how these relationships are made explicit.}

When considering plays as instances, no relationships between individual instances are made explicit. 
When considering characters or parts of plays as instances, however, relationships between characters, or between characters and parts of plays are made explicit in the graph representations, 
exploiting the TEI Simple encoding of that data and the annotations provided in the XML attributes.

\paragraph{Are there recommended data splits (e.g., training, development/validation, testing)?} 
\emph{If so, please provide a description of these splits, explaining the rationale behind them.}

There are no recommended data splits for the current
release.

\paragraph{Are there any errors, sources of noise, or redundancies in the dataset?}
\emph{If so, please provide a description.}

The raw data contain some errors and redundancies in the XML encoding. 
Errors include redundant XML tags (e.g., doubly-wrapped <div> tags), 
but also character entries or exits not explicitly annotated.
Redundancies result from the choice, made by the creators of \folgertexts, 
to encode some information conveyed in the raw text also as attributes or separate XML tags 
(e.g., a character who speaks is encoded both as an attribute of the tag wrapping the speech and as an XML tag wrapping the name of the speaker). 

There are two notable sources of noise affecting the preprocessed data and the graph data, 
both of which relate to our handling of stage directions---%
i.e., our processing of the XML attributes of \texttt{<stage>} tags in the raw data. 

First, to determine which characters are on stage when a word is spoken, 
we primarily rely on the contents of \texttt{who} attributes in the \texttt{<stage>} tags of the raw data marked with \texttt{type="entry"} resp. \texttt{type="exit"}. 
The \texttt{who} attributes, however, are sometimes \emph{semantically} incomplete, i.e., 
they may reflect Shakespeare's original stage directions accurately, 
but the original stage directions do not mention implied character movements (such as the exit of a side character or the exit of characters that died or fell unconscious at the end of a scene).
To limit the impact of this noise source on our graph representations, 
we ``flush'' characters when a new scene starts (to handle missing exits) 
and ensure that the speaker is always on stage (to handle missing entries, some of which are also introduced by our character flushing policy).

Second, in our directed graph representations, 
where edges encode speaking and being spoken to, 
we equate being on stage while a word is spoken with hearing the word.
Thus, we do not account for the impact of some stage directions concerning delivery, 
e.g., stage directions indicating that speech is inaudible for some or all other characters on stage, 
on the information flow our directed graph representations purport to capture. 
In the TEI simple encoding of our raw data, such stage directions are annotated with \texttt{type="delivery"}, 
but there is no indication of who can hear the words so delivered in the XML annotations. 
There are \checkednumber{2\thinspace 200} XML tags annotated with \texttt{type="delivery"} (i.e., \checkednumber{60} delivery modifications per play on average).
As modifications to delivery are sometimes crucial to drive the plot (e.g., by setting up misunderstandings), 
the impact of this noise source should not be underestimated, 
but it affects only our directed graph representations, 
which might be cautiously interpreted as ``upper bounds'' on the information flow between the characters on stage.

These sources of noise detailed above could likely be eliminated, to a large extent, by a more sophisticated parsing of the stage directions. 
This parsing could leverage, e.g., natural language processing methods to supplement the XML annotations. 
We plan to implement this improvement for a future dataset release.

\paragraph{Is the dataset self-contained, or does it link to or otherwise rely on external resources (e.g., websites, tweets, other datasets)?}
\emph{If it links to or relies on external resources, a) are there guarantees that they will exist, and remain constant, over time; b) are there official archival versions of the complete dataset (i.e., including the external resources as they existed at the time the dataset was created); c) are there any restrictions (e.g., licenses, fees) associated with any of the external resources that might apply to a dataset consumer? Please provide descriptions of all external resources and any restrictions associated with them, as well as links or other access points, as appropriate.}

The dataset is self-contained. 
The raw data stem from \folgertexts, maintained by the \folgerlibrary and released under the \folgerlicense license, and they are redistributed without modifications as part of the \ourdata dataset. 
All other data are derived from the raw data, and the \folgerlicense license does not impose any additional restrictions. 
As part of our dataset maintenance (see below), we will regularly check \folgertexts for modifications, 
and we will recompute and redistribute an updated \ourdata dataset under
a versioned DOI whenever we detect changes.

\paragraph{Does the dataset contain data that might be considered confidential (e.g., data that is protected by legal privilege or by doctor-patient confidentiality, data that includes the content of individuals' non-public communications)?} 
\emph{If so, please provide a description.}

The dataset does not contain data that might be considered confidential.

\paragraph{Does the dataset contain data that, if viewed directly, might be offensive, insulting, threatening, or might otherwise cause anxiety?} 
\emph{If so, please describe why.}

The raw data, i.e., Shakespeare's plays, contain scenes that might be considered offensive, insulting, threatening, or otherwise anxiety-inducing from a contemporary perspective.
For example, there is considerable controversy in the humanities around whether \emph{The Taming of the Shrew} is misogynistic, 
and the main female protagonist's final speech on female submissiveness (Act V, Scene 2, ll.~136--179) might cause discomfort to modern readers. 
Moreover, the corpus uses words that might be considered derogatory or offensive from a contemporary perspective.
The preprocessed data, however, disassembles the original text, such that (offensive) play content is no longer immediately apparent when the data is viewed directly.

%

\paragraph{Any other comments?}
~

The entire dataset takes up roughly 365 MB when uncompressed, and 30 MB when compressed.

\subsubsection{Collection Process}

\paragraph{How was the data associated with each instance acquired?} 
\emph{Was the data directly observable (e.g., raw text, movie ratings), reported by subjects (e.g., survey responses), or indirectly inferred/derived from other data (e.g., part-of-speech tags, model-based guesses for age or language)?}

The raw data associated with each instance was acquired from \folgertexts as XML files encoded in TEI Simple format. 
This format contains both raw text and structural, linguistic, and semantic annotations embedded in XML tags or XML attributes. 
Hence, it was partially directly observable (e.g., the raw text and its structure) 
and partially derived from other data (e.g., the XML tags and their attributes). 
The preprocessed data and the graph data were derived from the raw data.

\paragraph{If the data was reported by subjects or indirectly inferred/derived from other data, was the data validated/verified?} 
\emph{If so, please describe how.}

To the extent that the raw data were indirectly inferred or derived from other data, 
validation was performed by the specialists from \folgertexts.
The preprocessed data and the graph data were validated by unit tests and manual inspection aided by visualizations (which also led us to discover the noise sources detailed above).

\paragraph{What mechanisms or procedures were used to collect the data (e.g., hardware apparatuses or sensors, manual human curation, software programs, software APIs)?}
\emph{How were these mechanisms or procedures validated?}

The raw data was bulk downloaded in TEI Simple format as a ZIP archive from the \href{https://shakespeare.folger.edu/download-the-folger-shakespeare-complete-set/}{\folgertexts downloads section}, and \folgertexts compiled the raw data through computer-assisted manual curation.
The bulk download was checked manually to ensure that the extracted archive contained one XML file per play, as expected. 
The code creating the preprocessed data from the raw data and the graph representations from the preprocessed data is almost completely unit tested.

\paragraph{If the dataset is a sample from a larger set, what was the sampling strategy (e.g., deterministic, probabilistic with specific sampling probabilities)?}
~

The data is not a sample from a larger set.

\paragraph{Who was involved in the data collection process (e.g., students, crowdworkers, contractors) and how were they compensated (e.g., how much were crowdworkers paid)?}
~

Only \cc and \br, the dataset authors, were involved in the data collection process.

\paragraph{Over what timeframe was the data collected?} \emph{Does this timeframe match the creation timeframe of the data associated with the instances (e.g., recent crawl of old news articles)? If not, please describe the timeframe in which the data associated with the instances was created.}

The raw data was collected through one download call to\\
\href{https://shakespeare.folger.edu/downloads/teisimple/shakespeares-works_TEIsimple_FolgerShakespeare.zip}{https://shakespeare.folger.edu/downloads/teisimple/shakespeares-works\_TEIsimple\_FolgerShakespeare.zip}
in June 2022, 
and the preprocessed data and the graph data were derived from the raw data by running a code pipeline, also in June 2022.
This timeframe does not match the creation timeframe of the raw data, which, though internal to the \folgerlibrary, spans at least several months in 2020.
It also does not match the creation timeframe of Shakespeare's plays, which spans several decades in the 16th and 17th centuries.

\paragraph{Were any ethical review processes conducted (e.g., by an institutional review board)?} 
\emph{If so, please provide a description of these review processes, including the outcomes, as well as a link or other access point to any supporting documentation.}

No ethical review processes were conducted.

%

\paragraph{Any other comments?}
~

None.

\subsubsection{Preprocessing/Cleaning/Labeling}

\paragraph{Was any preprocessing/cleaning/labeling of the data done (e.g., discretization or bucketing, tokenization, part-of-speech tagging, SIFT feature extraction, removal of instances, processing of missing values)?} 
\emph{If so, please provide a description. If not, you may skip the remaining questions in this section.}

Our data preprocessing consists of two steps.
\begin{enumerate}
	\item Transform raw XML data into preprocessed CSV data (\texttt{rawdata}$\rightarrow$\texttt{data}).\\
	Script: \texttt{run\_preprocessing.py}
	\begin{enumerate}
		\item Extract the cast list from the TEI Simple XML and store it as a CSV. (This is technically unnecessary to generate our graph representations, but it gives a convenient overview of the characters occurring in the play.)\\
		Function: \texttt{get\_cast\_df}\\
		Artifact: \texttt{data/\{play\}.cast.csv}
		\item Parse the TEI Simple XML into a table containing one row per descendant of the TEI Simple \texttt{<body>} tag, 
		and the tag names and XML attributes of all XML tags of interest (eliminating redundant XML elements), 
		plus the text content of all XML tags that are leaves, as columns.
		Annotate the result with information on the act and scene in which the tag occurs, the characters on stage when the tag occurs, and the speaker(s), if any.\\
		Function \texttt{get\_raw\_xml\_df}\\
		Artifact: \texttt{data/\{play\}.raw.csv}
		\item Transform the artifact from the previous step into a table with one row per setting on stage, where a setting is a stretch of the play without changes to the speaker or to the group of characters on stage, 
		and information on the setting as well as the number of lines and tokens spoken in that setting as columns.\\
		Artifact: \texttt{data/\{play\}.agg.csv}
	\end{enumerate}
	\item Transform preprocessed CSV data into node and edge CSV files for graph construction (\texttt{data}$\rightarrow$\texttt{graphdata}).\\ 
	The artifacts resulting from this step are generally labeled \texttt{%
		\{play\}\_\{semantic mapping\}\_\{granularity\}\_\{expressivity\}.\{list type\}.csv}, 
	omitting the expressivity (and granularity) components in node lists if all different graph representations with a given semantic mapping (and granularity) use the same set of nodes.\\
	\begin{enumerate}
		\item Create node lists and edge lists for different graph representations in CSV format from \texttt{data/\{play\}.agg.csv} artifacts.\\ 
		 Script: \texttt{create\_graph\_representations.py}\\
		 Artifacts:
		 \begin{itemize}[label=--]
		 	\item \texttt{graphdata/\{play\}\_ce-group-mw.edges.csv}
		 	\item \texttt{graphdata/\{play\}\_ce-group-w.edges.csv}
		 	\item \texttt{graphdata/\{play\}\_ce-scene-mw.edges.csv}
		 	\item \texttt{graphdata/\{play\}\_ce-scene-w.edges.csv}
		 	\item \texttt{graphdata/\{play\}\_ce.nodes.csv}
		 	\item \texttt{graphdata/\{play\}\_se-group-w.edges.csv}
		 	\item \texttt{graphdata/\{play\}\_se-group.nodes.csv}
		 	\item \texttt{graphdata/\{play\}\_se-scene-w.edges.csv}
		 	\item \texttt{graphdata/\{play\}\_se-scene.nodes.csv}
		 	\item \texttt{graphdata/\{play\}\_se-speech-mwd.edges.csv}
		 	\item \texttt{graphdata/\{play\}\_se-speech-wd.edges.csv}
		 	\item \texttt{graphdata/\{play\}\_se-speech.nodes.csv}
		 \end{itemize}
		\item Create node lists and edge lists for different hypergraph representations in CSV format from \texttt{data/\{play\}.agg.csv} artifacts.\\ 
		Script: \texttt{create\_hypergraph\_representations.py}\\
		Artifacts: 
		\begin{itemize}[label=--]
			\item \texttt{graphdata/\{play\}\_hg-group-mw.edges.csv}
			\item \texttt{graphdata/\{play\}\_hg-group-mw.node-weights.csv}
			\item \texttt{graphdata/\{play\}\_hg-scene-mw.edges.csv}
			\item \texttt{graphdata/\{play\}\_hg-scene-mw.node-weights.csv}
			\item \texttt{graphdata/\{play\}\_hg-speech-mwd.edges.csv}
			\item \texttt{graphdata/\{play\}\_hg-speech-wd.edges.csv}
			\item \texttt{graphdata/\{play\}\_hg.nodes.csv}
		\end{itemize}
	\end{enumerate}
\end{enumerate}

\paragraph{Was the ``raw'' data saved in addition to the preprocessed/cleaned/labeled data (e.g., to support unanticipated future uses)?} 
\emph{If so, please provide a link or other access point to the “raw” data.}

The raw data was saved, and it is distributed along with the preprocessed data in the dataset available from \datarepo under a versioned DOI: \ourdatadoiall.

\paragraph{Is the software that was used to preprocess/clean/label the data available?} 
\emph{If so, please provide a link or other access point.}

The software used to transform the raw data into the preprocessed data, 
and the preprocessed data into the graph data representations, 
is available on \coderepo in the following repository:
\coderepourl.

All code releases are also available on \datarepo under a versioned DOI: \ourcodedoiall.

\paragraph{Any other comments?}
~

All data preprocessing can be completed in a couple of minutes even on older commodity hardware. 
We used a 2016 MacBook Pro with a 2.9 GHz Quad-Core Intel Core i7 processor and 16 GB RAM. 

\subsubsection{Uses}

\paragraph{Has the dataset been used for any tasks already?}
\emph{If so, please provide a description.}

In the paper introducing \ourdata, 
the dataset has been used to demonstrate the differences between rankings of characters by degree that result from different modeling choices made when transforming raw data into graphs.

\paragraph{Is there a repository that links to any or all papers or systems that use the dataset?}
\emph{If so, please provide a link or other access point.}

Papers or systems known to use dataset will be collected on \datapage and on \coderepo.

\paragraph{What (other) tasks could the dataset be used for?}
~

\ourdata was designed for inquiries into the stability of algorithmic results under different reasonable representations of the underlying raw data, 
i.e., to enable \emph{representation robustness checks} for \milena methods. 
In this role, it could generally be used for all \milena tasks identified as \emph{in scope} in the motivation section. 

\paragraph{Is there anything about the composition of the dataset or the way it was collected and preprocessed/cleaned/labeled that might impact future uses?} 
\emph{For example, is there anything that a dataset consumer might need to know to avoid uses that could result in unfair treatment of individuals or groups (e.g., stereotyping, quality of service issues) or other risks or harms (e.g., legal risks, financial harms)? If so, please provide a description. Is there anything a dataset consumer could do to mitigate these risks or harms?}

The quality and expressivity of the dataset is limited by the quality and expressivity of \folgertexts encoded using the TEI Simple format, 
which could restrict usage in the digital humanities, 
e.g., when they are interested in the minute details of character interactions described in stage directions. 

\ourdata contains relational data representations of Shakespeare's plays, 
which were written more than four centuries ago. 
Hence, there are no risks or harms associated with the dataset 
beyond the risks or harms also associated with the ongoing study of Shakespeare's works in the humanities, 
and the risks or harms associated with the decontextualization or overinterpretation of any dataset.

At \datapage and on \coderepo, we keep a continuously-updated list of all known dataset limitations for dataset consumers to review when deciding whether \ourdata is appropriate for their use case.

\paragraph{Are there tasks for which the dataset should not be used?}
\emph{If so, please provide a description.}
~

Outside \emph{representation robustness checks}, 
\ourdata should not be used in tasks that have no reasonable semantic interpretation in the domain of the raw data.

\paragraph{Any other comments?}
~

None.

\subsubsection{Distribution}\label{subsec:distribution}

\paragraph{Will the dataset be distributed to third parties outside of the entity (e.g., company, institution, organization) on behalf of which the dataset was created?}
\emph{If so, please provide a description.}

The dataset was not created on behalf of any entity, and it will be distributed freely. 

\paragraph{How will the dataset will be distributed (e.g., tarball on website, API, GitHub)?}
\emph{Does the dataset have a digital object identifier (DOI)?}

The dataset will be distributed as a ZIP archive via \datarepo, based on code hosted on \coderepo. 
Each dataset version and each code release will have a versioned DOI, 
generated automatically by \datarepo.
See also Table~\ref{tab:access}.

\paragraph{When will the dataset be distributed?}
~

The dataset will be distributed when the paper introducing it is submitted.

\paragraph{Will the dataset be distributed under a copyright or other intellectual property (IP) license, and/or under applicable terms of use (ToU)?}
\emph{If so, please describe this license and/or ToU, and provide a link or other access point to, or otherwise reproduce, any relevant licensing terms or ToU, as well as any fees associated with these restrictions.}

The dataset will be distributed under a \ourdatalicense license, according to which others are free to
\begin{itemize}[label=--]
	\item \emph{share}, i.e., copy and redistribute, and
	\item \emph{adapt}, i.e., remix, transform, and build on the material,
\end{itemize}
provided they
\begin{itemize}[label=--]
	\item \emph{give attribution}, i.e., give appropriate credit, provide a link to the license, and indicate if changes were made,
	\item do \emph{not} use the material for \emph{commercial purposes}, and
	\item \emph{add no restrictions} limiting others in doing anything the license permits.
\end{itemize}

The code constructing the dataset will be distributed under a permissive \ourcodelicense license.

\paragraph{Have any third parties imposed IP-based or other restrictions on the data associated with the instances?}
\emph{If so, please describe these restrictions, and provide a link or other access point to, or otherwise reproduce, any relevant licensing terms, as well as any fees associated with these restrictions.}

The \folgerlibrary has released the source of our raw data, \folgertexts, under the \folgerlicense license, 
which has essentially the same usage conditions as our \ourdatalicense license.

\paragraph{Do any export controls or other regulatory restrictions apply to the dataset or to individual instances?}
\emph{If so, please describe these restrictions, and provide a link or other access point to, or otherwise reproduce, any supporting documentation.}

No export controls or other regulatory restrictions apply.

\paragraph{Any other comments?}
~

None.

\subsubsection{Maintenance}
\label{subsec:maintenance}

\paragraph{Who will be supporting/hosting/maintaining the dataset?}
~

\cc and \br will be supporting, hosting, and maintaining the dataset. 

\paragraph{How can the owner/curator/manager of the dataset be contacted (e.g., email address)?}
~

In the interest of transparency, the preferred method to contact the dataset maintainers is by opening \coderepo issues at \coderepourl. 
Alternatively, the dataset maintainers can be reached by email to \href{mailto:info@hyperbard.net}{info@hyperbard.net}

\paragraph{Is there an erratum?}
\emph{If so, please provide a link or other access point.}

Errata will be documented at \datapage and on \coderepo.

\paragraph{Will the dataset be updated (e.g., to correct labeling errors, add new instances, delete instances)?}
\emph{If so, please describe how often, by whom, and how updates will be communicated to dataset consumers (e.g., mailing list, GitHub)?}

The dataset will be updated as needed, and updates will be labeled using \emph{semantic versioning}. 
\begin{itemize}[label=--]
	\item A \emph{patch version} (e.g., 0.0.1 $\rightarrow$ 0.0.2) is a recomputation of the latest dataset version following a non-breaking change in the underlying raw data. 
	\item A \emph{minor version} (e.g., 0.0.1 $\rightarrow$ 0.2.0) is an update of the latest dataset version that increases the expressivity of existing representations while maintaining all of their previously present features.
	\item Any other update is a \emph{major version} (e.g., 0.0.1 $\rightarrow$ 1.0.0). 
	This includes, e.g., responses to breaking changes in the underlying source data, 
	additions of new representations, 
	and changes to existing representations that might break dataset consumer code. 
\end{itemize}
Patch versions will be created automatically using \coderepo actions.
Minor versions and major versions will be created by the dataset maintainers, 
potentially accepting pull requests or implementing feature requests filed via at \coderepourl.

New releases will be communicated at \datapage and on \coderepo, 
and they will be available for download under a versioned DOI on \datarepo, with \ourdatadoiall always resolving to the latest release.

\paragraph{If the dataset relates to people, are there applicable limits on the retention of the data associated with the instances (e.g., were the individuals in question told that their data would be retained for a fixed period of time and then deleted)?}
\emph{If so, please describe these limits and explain how they will be enforced.}

There are no data retention limits.

\paragraph{Will older versions of the dataset continue to be supported/hosted/maintained?}
\emph{If so, please describe how. If not, please describe how its obsolescence will be communicated to dataset consumers.}

Older versions of the dataset will remain hosted on \datarepo, 
with the relevant version of the code needed to reproduce them available in an associated \coderepo release, 
also archived on \datarepo.

There will be basic support for older versions of the dataset, 
and as \ourdata is derived from century-old literary works, 
dataset maintenance amounts to dataset updates (see the paragraph on dataset updates).

\paragraph{If others want to extend/augment/build on/contribute to the dataset, is there a mechanism for them to do so?}
\emph{If so, please provide a description. Will these contributions be validated/verified? If so, please describe how. If not, why not? Is there a process for communicating/distributing these contributions to dataset consumers? If so, please provide a description.}

Others can extend, augment, build on, and contribute to the dataset through the engagement mechanisms provided by \coderepo.\\
See also \href{https://github.com/hyperbard/hyperbard/blob/main/CONTRIBUTING.md}{https://github.com/hyperbard/hyperbard/blob/main/CONTRIBUTING.md}.

Extensions, augmentations, and contributions provided via pull requests will be validated and verified by the dataset maintainers in a regular code and data review process, 
while changes made in independent forks will not be checked.

Contributions integrated with the \ourdata code repository will be visible on \coderepo, and they trigger new dataset releases, in which contributors will be specifically acknowledged.

\paragraph{Any other comments?}
~

None.

\subsection{Hosting, License, and Maintenance Plan}

For hosting and licensing information, see Table~\ref{tab:access} and Section~\ref{subsec:distribution}. 
For the maintenance plan, see Section~\ref{subsec:maintenance}.

\subsection{Author Responsibility Statement}

The dataset authors, \cc and \br, bear all responsibility in case of violation of rights, etc., and they confirm that the data is released under the \ourdatalicense license, and that the code is released under the \ourcodelicense license.

%
%
%


\section{Usage Documentation}

The \ourdata dataset is distributed in four folders: 
\texttt{rawdata}, \texttt{data}, \texttt{graphdata}, and \texttt{metadata}. 
See Section~\ref{subsec:composition} for more details on the composition of the dataset.
The dataset can be reproduced by cloning the \coderepo repository and running \texttt{make} (this will also generate most figures included in the \ourdata paper).

In addition to the written documentation, we provide Jupyter notebook tutorials for interactive data exploration. 
The tutorials are hosted on \coderepo at
\tutorialrepourl,
and they can be run both locally and in a \binderurl, i.e., a fully configured remote environment accessible through the browser without any local setup.
Launching the \binderurl usually takes around thirty seconds.

In the following, we explain the structure of the files in \ourdata's folders 
and detail how these files can be read. 
All file examples are taken from \randj, 
and for CSV files, all columns are described in alphabetical order.

\subsection{\texttt{rawdata}}

This folder contains XML files encoded in TEI Simple as provided by \folgertexts. 
These files can be read with any XML parser, such as the parser from the \texttt{beautifulsoup4} library in Python. 
All file names follow the pattern \texttt{\{play\}\_TEIsimple\_FolgerShakespeare.xml}.

The XML encoding is designed to meet the needs of the (digital) humanities, 
and hence, it is very detailed and fine-grained. 
For example, every word, whitespace character, and punctuation mark is contained in its own tag.

The encoding practices followed by \folgertexts are described in the \texttt{<encodingDesc>} tag of each text. 
To summarize:
\begin{itemize}[label=--]
	\item The major goal of the TEI Simple encoding is to achieve interoperability with a large corpus of early modern texts derived from the Early English Books Text Creation Partnership transcriptions (i.e., it is different from \emph{our} goal).
	\item The encoding is completely faithful to the readings, orthography, and punctuation of the source texts (i.e., the Shakespeare texts edited by Barbara Mowat and Paul Werstine at \folgerlibrary).
	\item All \texttt{xml:id}s are corpuswide identifiers (i.e., they are unique across all our plays, too).
	\item Words, spaces, and punctuation characters are numbered sequentially within each play, incremented by 10 (XML attribute: \texttt{n}).
	\item Most other elements begin with an element-specific prefix, followed by a reference to the Folger Through Line Number, a sequential numbering of the numbered lines in the text. (Details omitted.)
	\item Spoken words are linguistically annotated with a lemma and POS tag.
\end{itemize}

Running the script \texttt{compute\_rawdata\_xml\_statistics.py} in the \ourdata \coderepo repository, 
which computes basic XML tag, path, and attribute statistics for the entire corpus and writes the results to the \texttt{metadata} folder as CSV files, 
provides some intuition regarding the structure of the raw data. 
This script also pulls the descriptions of all tags from the current TEI specification.
For more information on the TEI Simple format, which has been integrated with the main TEI specification, see \href{https://github.com/TEIC/TEI-Simple}{https://github.com/TEIC/TEI-Simple}.

Example:
\begin{verbatim}
	...
	<sp xml:id="sp-0015" who="#SERVANTS.CAPULET.Sampson_Rom">
	<speaker xml:id="spk-0015">
	<w xml:id="fs-rom-0002610">SAMPSON</w>
	</speaker>
	<p xml:id="p-0015">
	<lb xml:id="ftln-0015" n="1.1.1"/>
	<w xml:id="fs-rom-0002620" n="1.1.1" lemma="Gregory" ana="#n1-nn">Gregory</w>
	<pc xml:id="fs-rom-0002630" n="1.1.1">,</pc>
	<c> </c>
	<w xml:id="fs-rom-0002650" n="1.1.1" lemma="on" ana="#acp-p">on</w>
	<c> </c>
	<w xml:id="fs-rom-0002670" n="1.1.1" lemma="my" ana="#po">my</w>
	<c> </c>
	<w xml:id="fs-rom-0002690" n="1.1.1" lemma="word" ana="#n1">word</w>
	<c> </c>
	<w xml:id="fs-rom-0002710" n="1.1.1" lemma="we|will" ana="#pns|vmb">we’ll</w>
	<c> </c>
	<w xml:id="fs-rom-0002730" n="1.1.1" lemma="not" ana="#xx">not</w>
	<c> </c>
	<w xml:id="fs-rom-0002750" n="1.1.1" lemma="carry" ana="#vvi">carry</w>
	<c> </c>
	<w xml:id="fs-rom-0002770" n="1.1.1" lemma="coal" ana="#n2">coals</w>
	<pc xml:id="fs-rom-0002780" n="1.1.1">.</pc>
	</p>
	</sp>
	...
\end{verbatim}

\subsection{\texttt{data}}

This folder contains CSV files, which can be read with any CSV parser, such as the parser from the \texttt{pandas} library in Python.

There are three types of files:\\ 
\texttt{\{play\}.cast.csv} files, \texttt{\{play\}.raw.csv} files, and \texttt{\{play\}.agg.csv} files.

\subsubsection{\texttt{\{play\}.cast.csv}}

A \texttt{\{play\}.cast.csv} file contains the XML identifiers and attributes of all \texttt{<castItem>} tags found in a \texttt{\{play\}\_TEIsimple\_FolgerShakespeare.xml} file. 
It gives an overview of the characters occurring in a play, 
and it can be used to count the number of characters (including characters that do not speak) or to build a hierarchy of characters and character groups.

Rows correspond to characters or character groups.

Columns in alphabetical order:
\begin{itemize}[label=--]
\item \texttt{corresp}: group (i.e., another cast item) to which a given cast item belongs, if any (XML attribute abbreviating ``corresponds'').\\ 
Type: String or NaN (if the cast item does not belong to any other cast item).
\item \texttt{xml:id}: 	unique identifier of the cast member.\\
Type: String.
\end{itemize}
Note that the data in each of these columns does \emph{not} start with a \# sign. 
This contrasts with \emph{references} to the \texttt{xml:id}s in the attributes of other XML tags in the raw data XML files, which \emph{do} start with a \# sign (to indicate the referencing).

Example:
\begin{lstlisting}
	xml:id,corresp
	ATTENDANTS.PRINCE_Rom,ATTENDANTS_Rom
	ATTENDANTS_Rom,
	Apothecary_Rom,
	Benvolio_Rom,
	Boy_Rom,
	...
\end{lstlisting}

\subsubsection{\texttt{\{play\}.raw.csv}}

A \texttt{\{play\}.raw.csv} file contains the descendants of the \texttt{<body>} tag found in a \texttt{\{play\}\_TEIsimple\_FolgerShakespeare.xml} file, 
with redundancies resulting from the encoding format eliminated, and additional information to build graph representations annotated. 
It provides a \emph{disaggregated} tabular overview of the information underlying our graph representations, 
and it serves as the basis of its corresponding \texttt{\{play\}.agg.csv} file.

Rows correspond to instances of XML tags.

Columns in alphabetical order:
\begin{itemize}[label=--]
\item \texttt{act}: Derived attribute. The number of the act in which the tag occurs. 
An integer in $[5]$ for all tags in the main part of the play. 
$0$ for tags occurring before the first act (e.g., in a prologue or an induction), $6$ for tags occurring after the fifth act (e.g., in an epilogue).\\ 
Type: Non-negative integer.
\item \texttt{ana}: Original attribute. If the tag wraps a spoken word, the POS tag of that word (XML attribute abbreviating ``analysis'').\\ 
Type: String or NaN (if the tag does not wrap a spoken word).
\item \texttt{lemma}: Original attribute. If the tag wraps a spoken word, the lemma of that word.\\
Type: String or NaN (if the tag does not wrap a spoken word).
\item \texttt{n}: Original attribute. A label for the element, not necessarily unique.\\
Type: String, positive integer (for \texttt{<div>} tags representing acts or scenes), or NaN (e.g., for \texttt{<c>} tags wrapping whitespace characters).
\item \texttt{onstage}: Derived attribute. Whitespace-separated list of characters on stage when the tag occurs.\\
Type: String or NaN.
\item \texttt{part}: Original attribute. Rare and not of interest for graph building.\\
Type: String or NaN.
\item \texttt{prev}: Original attribute. Rare and not of interest for graph building.\\
Type: String or NaN.
\item \texttt{rendition}: Original attribute. Rare and not of interest for graph building.\\
Type: String or NaN.
\item \texttt{scene}: Derived attribute. 
The number of the scene in which the tag occurs. 
$0$ if the tag does not occur in a scene.\\
Type: Non-negative integer.
\item \texttt{speaker}: Derived attribute. Whitespace-separated list of characters who are speaking when a tag occurs. 
Note that several characters can speak at the same time, although the overwhelming majority of speech in the corpus is uttered by only one speaker.\\
Type: String or NaN.
\item \texttt{stagegroup\_raw}: Derived attribute. Number stating how many changes in the set of characters on stage we have already witnessed when a tag occurs (i.e., the same set of characters can occur in different stage groups). Relevant for sorting and aggregation.\\
Type: Non-negative integer.
\item \texttt{tag}: Original entity. The name of the XML tag to which the row corresponds.\\
Type: String.
\item \texttt{text}: Original text content.\\
Type: String or NaN (if a tag is not a leaf in the XML tree).
\item \texttt{type}: Original attribute. Used to give details on \texttt{<div>} and \texttt{<stage>} tags, e.g., distinguish between acts and scenes, and mark stage directions as, e.g., character entry or exit.\\
Type: String or NaN.
\item \texttt{who}: Original attribute giving information on characters who act, transformed into a set. Will become whitespace-separated list in future releases.\\
Type: Set of strings or NaN. 
\item \texttt{xml:id}: Original XML identifier. Note that instances of some XML tags, including \texttt{<div>} and \texttt{<c>} tags, do not have XML identifiers.\\
Type: String or NaN.
\end{itemize}

Example:
\begin{lstlisting}
	tag,type,n,text,xml:id,who,lemma,ana,part,rendition,prev,act,scene,onstage,stagegroup_raw,speaker
	...
	sp,,,,sp-0015,{'#SERVANTS.CAPULET.Sampson_Rom'},,,,,,1,1,#SERVANTS.CAPULET.Gregory_Rom #SERVANTS.CAPULET.Sampson_Rom,3,#SERVANTS.CAPULET.Sampson_Rom
	p,,,,p-0015,,,,,,,1,1,#SERVANTS.CAPULET.Gregory_Rom #SERVANTS.CAPULET.Sampson_Rom,3,
	lb,,1.1.1,,ftln-0015,,,,,,,1,1,#SERVANTS.CAPULET.Gregory_Rom #SERVANTS.CAPULET.Sampson_Rom,3,#SERVANTS.CAPULET.Sampson_Rom
	w,,1.1.1,Gregory,fs-rom-0002620,,Gregory,#n1-nn,,,,1,1,#SERVANTS.CAPULET.Gregory_Rom #SERVANTS.CAPULET.Sampson_Rom,3,#SERVANTS.CAPULET.Sampson_Rom
	pc,,1.1.1,",",fs-rom-0002630,,,,,,,1,1,#SERVANTS.CAPULET.Gregory_Rom #SERVANTS.CAPULET.Sampson_Rom,3,#SERVANTS.CAPULET.Sampson_Rom
	c,,, ,,,,,,,,1,1,#SERVANTS.CAPULET.Gregory_Rom #SERVANTS.CAPULET.Sampson_Rom,3,
	w,,1.1.1,on,fs-rom-0002650,,on,#acp-p,,,,1,1,#SERVANTS.CAPULET.Gregory_Rom #SERVANTS.CAPULET.Sampson_Rom,3,#SERVANTS.CAPULET.Sampson_Rom
	c,,, ,,,,,,,,1,1,#SERVANTS.CAPULET.Gregory_Rom #SERVANTS.CAPULET.Sampson_Rom,3,
	...
\end{lstlisting}

\subsubsection{\texttt{\{play\}.agg.csv}}

A \texttt{\{play\}.agg.csv} file contains a condensed and filtered view of its corresponding \texttt{\{play\}.raw.csv} file, focusing only on spoken words.
It provides an \emph{aggregated} tabular overview of the information underlying our graph representations, and it serves as the basis of all files in the \texttt{graphdata} folder. 
In contrast to the \texttt{\{play\}.raw.csv} file, which contains some original attributes, 
\texttt{\{play\}.agg.csv} contains only derived attributes.

Rows correspond to \emph{settings} (or \emph{speech acts}), i.e., maximal sequences of words in which neither the speaker(s) nor the group of characters on stage change.

Columns in alphabetical order:
\begin{itemize}[label=--]
\item \texttt{act}: The same as \texttt{act} in \texttt{\{play\}.raw.csv}.
\item \texttt{n\_lines}: The number of lines spoken in a setting.\\
Type: Positive integer.
\item \texttt{n\_tokens}: The number of tokens spoken in a setting.\\
Type: Positive integer.
\item \texttt{onstage}: The same as \texttt{onstage} in \texttt{\{play\}.raw.csv}.
\item \texttt{scene}: The same as \texttt{scene} in \texttt{\{play\}.raw.csv}.
\item \texttt{setting}: Number stating how many changes in the tuple (set of characters on stage, speaker) we have seen when the words summarized in this row occur, plus $1$ (for consistency with the numbering in \texttt{stagegroup}).\\
Type: Positive integer.
\item \texttt{speaker}: The same as \texttt{speaker} in \texttt{\{play\}.raw.csv}.
\item \texttt{stagegroup}: The contents of the \texttt{stagegroup\_raw} column, renumbered to be consecutive in \texttt{\{play\}.agg.csv}, starting with $1$.\\
Type: Positive integer.
\item \texttt{stagegroup\_raw}: The same as \texttt{stagegroup\_raw} in \texttt{\{play\}.raw.csv}.
\end{itemize}

Example:
\begin{lstlisting}
	act,scene,stagegroup,stagegroup_raw,setting,onstage,speaker,n_lines,n_tokens
	0,0,1,1,1,#Chorus_Rom,#Chorus_Rom,14,106
	1,1,2,3,2,#SERVANTS.CAPULET.Gregory_Rom #SERVANTS.CAPULET.Sampson_Rom,#SERVANTS.CAPULET.Sampson_Rom,1,8
	1,1,2,3,3,#SERVANTS.CAPULET.Gregory_Rom #SERVANTS.CAPULET.Sampson_Rom,#SERVANTS.CAPULET.Gregory_Rom,1,7
	1,1,2,3,4,#SERVANTS.CAPULET.Gregory_Rom #SERVANTS.CAPULET.Sampson_Rom,#SERVANTS.CAPULET.Sampson_Rom,1,9
	1,1,2,3,5,#SERVANTS.CAPULET.Gregory_Rom #SERVANTS.CAPULET.Sampson_Rom,#SERVANTS.CAPULET.Gregory_Rom,2,10
\end{lstlisting}

\subsection{\texttt{graphdata}}

This folder contains CSV files, which can be read with any CSV parser, such as the parser from the \texttt{pandas} library in Python.

For each play, the folder holds all files needed to generate the representations listed in Table~\ref{tab:representations}, i.e.:
\begin{itemize}[label=--]
\item Files to construct \emph{clique expansions} (\texttt{ce}, i.e., character co-occurrence networks):
\begin{itemize}[label=--]
	\item \texttt{\{play\}\_ce-group-mw.edges.csv}:\\ 
	Weighted multi-edges for clique expansions aggregated at the stage group level.\\
	Use to generate ce-group-\{mb,mw\} representations.
	\item \texttt{\{play\}\_ce-group-w.edges.csv}:\\
	Count-weighted edges for clique expansions aggregated at the stage group level.\\
	Use to generate ce-group-b representations (or ce-group-w representations for easier plotting of ce-group-mb representations if the edge order does not matter).
	\item \texttt{\{play\}\_ce-scene-mw.edges.csv}:\\
	Weighted multi-edges for clique expansions aggregated at the scene level.\\
	Use to generate ce-scene-\{mb,mw\} representations.
	\item \texttt{\{play\}\_ce-scene-w.edges.csv}:\\
	Count-weighted edges for clique expansions aggregated at the scene level.\\
	Use to generate ce-scene-b representations (or ce-scene-w representations for easier plotting of ce-scene-mb representations if the edge order does not matter).
	\item \texttt{\{play\}\_ce.nodes.csv}:\\ 
	Nodes for all clique expansions.\\
	Use to generate all ce-$*$ representations.
\end{itemize}
\item Files to construct \emph{star expansions} (\texttt{se}, i.e., bipartite graphs with characters and text units as node sets):
\begin{itemize}[label=--]
	\item \texttt{\{play\}\_se-group-w.edges.csv}:\\
	Edges for star expansions aggregated at the stage group level.\\ 
	Use to generate se-group-\{b,w\} representations.
	\item \texttt{\{play\}\_se-group.nodes.csv}:\\
	Nodes for star expansions aggregated at the stage group level.\\
	Use to generate se-group-\{b,w\} representations. 
	\item \texttt{\{play\}\_se-scene-w.edges.csv}:\\
	Edges for star expansions aggregated at the scene level.\\
	Use to generate se-scene-\{b,w\} representations.
	\item \texttt{\{play\}\_se-scene.nodes.csv}:\\
	Nodes for star expansions aggregated at the scene level. 
	The character nodes are the same as for \texttt{\{play\}\_se-group.nodes.csv}, but the text unit nodes differ.\\
	Use to generate se-scene-\{b,w\} representations.
	\item \texttt{\{play\}\_se-speech-mwd.edges.csv}:\\
	Directed multi-edges for star expansions aggregated at the speech act level. 
	Multi-edges can occur because there exists one edge per speech act, but text unit nodes are resolved at the stage group level, and one stage group can contain several speech acts.\\
	Use to generate the se-speech-mwd representation.
	\item \texttt{\{play\}\_se-speech-wd.edges.csv}:\\
	Directed edges for star expansions aggregated at the speech act level, with multi-edges aggregated into edge weights.\\
	Use to generate the se-speech-wd representation.
	\item \texttt{\{play\}\_se-speech.nodes.csv}:\\ 
	Nodes for star expansions aggregated at the speech act level. 
	The same as \texttt{\{play\}\_se-group.nodes.csv}; provided separately to facilitate the matching between node and edge files.\\
	Use to generate se-speech-\{wd,mwd\} representations.
\end{itemize}
\item Files to construct \emph{hypergraphs} (\texttt{hg}, i.e., generalized graph representations allowing edges with cardinalities in $\mathbb{N}$):
\begin{itemize}[label=--]
	\item \texttt{\{play\}\_hg-group-mw.edges.csv}:\\
	Edges for hypergraph representations resolved at the stage group level.\\
	Use to generate hg-group-\{mb,mw\} representations.
	\item \texttt{\{play\}\_hg-group-mw.node-weights.csv}:\\
	Edge-specific node weights for hypergraph representations resolved at the stage group level.\\
	Use to generate hg-group-\{mb,mw\} representations with edge-specific node weights.
	\item \texttt{\{play\}\_hg-scene-mw.edges.csv}:\\
	Edges for hypergraph representations resolved at the scene level.\\
	Use to generate hg-scene-\{mb,mw\} representations.
	\item \texttt{\{play\}\_hg-scene-mw.node-weights.csv}:\\
	Edge-specific node weights for hypergraph representations resolved at the scene level.\\
	Use to generate hg-scene-\{mb,mw\} representations with edge-specific node weights.
	\item \texttt{\{play\}\_hg-speech-mwd.edges.csv}:\\
	Directed, weighted multi-edges for hypergraph representations resolved at the speech act level, where both the source and the target can contain multiple nodes.\\
	Use to generate the hg-speech-mwd representation.
	\item \texttt{\{play\}\_hg-speech-wd.edges.csv}:\\
	Directed, weighted edges for hypergraph representations resolved at the speech act level, where both the source and the target can contain multiple nodes, with multi-edges aggregated into edge weights\\
	Use to generate the hg-speech-wd representation.
	\item \texttt{\{play\}\_hg.nodes.csv}:\\ 
	Nodes for all hypergraph representations. Technically redundant because hyperedges can have cardinality $1$, too, such that all nodes can be derived from the edge lists. Provided with global node weights for convenience.\\
	Use to generate all hg-$*$ representations.
\end{itemize}
\end{itemize}

The rows in each file represent either nodes or edges.

The columns in the individual files differ depending on the \emph{semantic mapping}, the \emph{granularity}, and the \emph{expressivity} of the file contents, all of which are expressed in the file name (cf. Table~\ref{tab:representations}), 
but the column semantics should be intuitive in light of the details on the \texttt{\{play\}.agg.csv} file columns given above. 
Note the following conventions for column names in edge lists:
\begin{itemize}[label=--]
	\item For clique and star expansions, if the graph is undirected, the nodes are called \texttt{node1} and \texttt{node2}, and if the graph is directed, the nodes are called \texttt{source} and \texttt{target}.
	\item If edges are count-weighted, the weight column is called \texttt{count}, otherwise, the columns \texttt{n\_tokens} and \texttt{n\_lines} can both serve as edge weights.
	\item For multi-edges in clique and star expansions, the column \texttt{edge\_index} ensures that there are no duplicate rows. In hypergraphs, this is ensured by the \texttt{setting} column.
\end{itemize}
Finally, when working with the edge lists, please refer to the \emph{expressivity} column in Table~\ref{tab:representations} to check whether the edge ordering in any particular file is intrinsically meaningful.

Examples:
\begin{itemize}[label=--]
	\item Nodes for clique expansions:
	\begin{verbatim}
		node
		#ATTENDANTS.PRINCE_Rom
		#ATTENDANTS_Rom
		#Apothecary_Rom
		#Benvolio_Rom
		#Boy_Rom
		...
	\end{verbatim}
	\item Edges for clique expansions (here: ce-group-mw):
	\begin{verbatim}
		node1,node2,key,act,scene,stagegroup,n_tokens,n_lines,edge_index
		#SERVANTS.CAPULET.Gregory_Rom,#SERVANTS.CAPULET.Sampson_Rom,0,1,1,2,254,33,2
		#SERVANTS.CAPULET.Gregory_Rom,#SERVANTS.CAPULET.Sampson_Rom,1,1,1,3,149,25,3
		#SERVANTS.CAPULET.Gregory_Rom,#SERVANTS.MONTAGUE.1_Rom,0,1,1,3,149,25,3
		#SERVANTS.CAPULET.Gregory_Rom,#SERVANTS.MONTAGUE.Abram_Rom,0,1,1,3,149,25,3
		#SERVANTS.CAPULET.Sampson_Rom,#SERVANTS.MONTAGUE.1_Rom,0,1,1,3,149,25,3
		...
	\end{verbatim}
	\item Nodes for star expansions (here: se-group):
	\begin{verbatim}
		node,node_type
		#ATTENDANTS.PRINCE_Rom,character
		#ATTENDANTS_Rom,character
		#Apothecary_Rom,character
		...
		0.00.0001,text_unit
		1.01.0002,text_unit
		1.01.0003,text_unit
		...
	\end{verbatim}
	\item Edges for star expansions (here: se-speech-mwd):
	\begin{verbatim}
		source,target,key,n_lines,n_tokens,edge_index,edge_type
		#Chorus_Rom,0.00.0001,0,14,106,1,active
		#SERVANTS.CAPULET.Sampson_Rom,1.01.0002,0,1,8,2,active
		1.01.0002,#SERVANTS.CAPULET.Gregory_Rom,0,1,8,2,passive
		#SERVANTS.CAPULET.Gregory_Rom,1.01.0002,0,1,7,3,active
		1.01.0002,#SERVANTS.CAPULET.Sampson_Rom,0,1,7,3,passive
		...
	\end{verbatim}
	\item Nodes for hypergraphs: 
	\begin{verbatim}
		node,n_tokens_onstage,n_tokens_speaker,n_lines_onstage,n_lines_speaker
		#ATTENDANTS.PRINCE_Rom,1147,0,150,0
		#ATTENDANTS_Rom,905,0,121,0
		#Apothecary_Rom,224,53,29,7
		#Benvolio_Rom,5671,1160,771,161
		#Boy_Rom,905,0,121,0
		...
	\end{verbatim}
	\item Edge-specific node weights for hypergraphs (here: hg-scene-mw):
	\begin{verbatim}
		act,scene,node,n_tokens_speaker,n_lines_speaker,n_tokens_onstage,n_lines_onstage
		0,0,#Chorus_Rom,106,14,106,14
		1,1,#Benvolio_Rom,376,52,1403,189
		1,1,#CITIZENS_Rom,16,2,237,32
		1,1,#Capulet_Rom,26,3,221,30
		1,1,#LadyCapulet_Rom,10,2,221,30
		...
	\end{verbatim}
	\item Edges for hypergraphs (here: hg-speech-mwd):
\end{itemize}
\begin{lstlisting}
	act,scene,stagegroup,setting,speaker,onstage,n_tokens,n_lines
	0,0,1,1,#Chorus_Rom,#Chorus_Rom,106,14
	1,1,2,2,#SERVANTS.CAPULET.Sampson_Rom,#SERVANTS.CAPULET.Gregory_Rom #SERVANTS.CAPULET.Sampson_Rom,8,1
	1,1,2,3,#SERVANTS.CAPULET.Gregory_Rom,#SERVANTS.CAPULET.Gregory_Rom #SERVANTS.CAPULET.Sampson_Rom,7,1
	1,1,2,4,#SERVANTS.CAPULET.Sampson_Rom,#SERVANTS.CAPULET.Gregory_Rom #SERVANTS.CAPULET.Sampson_Rom,9,1
	...
\end{lstlisting}

\subsection{\texttt{metadata}}

This folder currently contains exactly one CSV file, which maps play identifiers to play types. 
The file can be read with any CSV parser, such as the parser from the \texttt{pandas} library in Python, but since its provenance is documented as a comment at the start of the file, 
the \# character needs to be passed to the parser as a comment character.

Rows correspond to plays.

Columns in alphabetical order: 
\begin{itemize}[label=--]
	\item \texttt{play\_name}: The name of the play, as used to fill the \texttt{\{play\}} placeholder in all play-specific file names.\\
	Type: String.
	\item \texttt{play\_type}: The type of the play. One of $\{\text{comedy}, \text{history}, \text{tragedy}\}$.\\
	Type: String.
\end{itemize}

\section{Contribution Documentation}
\label{apx:contributions}

In the following, for context and accessibility, we summarize the story of the full play as well as its two main themes, the dataset and the community critique.

\subsection{The Story}
	\emph{Induction, Scene~I.}
	Confronted by \reviewer, \playauthors explain their first contribution.
	\emph{Act~I, Scene~I.}
	\creature gets drawn into the Community by \seniorresearcher and \tutor. 
	Welcomed by \professor, they sign their PhD contract.
	\emph{Act~I, Scene~II.}
	\creature quarrels with their new role. 
	They meet \colleague, their office mate, and three \deadlines, introduced by \professor.
	They submit to \deadlineone.
	\emph{Act~I, Scene~III.}
	\creature dreams of \hyperbard, a faun caring for raw data, and \graph, one of their spirits. 
	They discuss how to obtain insights from raw data via transformations, 
	and that each raw data point permits several relational representations.
	\emph{Act~II, Scene~I.}
	\creature converses with \colleague, \professor, and \seniorresearcher over lunch. 
	They ask \colleague about the provenance of graph data used in the Community, and they learn about graph data repositories.
	\emph{Act~II, Scene~II.}
	\creature revisits their dream. 
	They identify semantic mapping, granularity, and expressivity as the dimensions in which several graph representations of the same raw data may differ.
	\emph{Act~II, Scene~III.}
	\creature secretly observes \colleague as they mechanically prepare a graph dataset and produce a datasheet in the process.
	\emph{Act~II, Scene~IV.}
	Confused and depressed by the practices they witness in the Community, \creature attempts suicide.
	\emph{Act~II, Scene~V.}
	Outside the Community, \creature is cared for by \graph and \hyperbard.
	Together, the three of them develop the graph and hypergraph representations of Shakespeare's plays included in the \ourdata dataset.
	\emph{Act~III, Scene~I.}
	\creature gets haunted by the three \deadlines, who remind them of their
	ignoble academic incentives.
	They contemplate quitting their PhD.
	\emph{Act~IV, Scene~I.}
	Accompanied by \graph and \hyperbard, \creature returns to the Community. 
	They meet \professor, who calls \creature into their office and demands that \hyperbard leaves.
	\emph{Act~IV, Scene~II.}
	From \professor, \creature learns that their paper got accepted.
	\emph{Act~IV, Scene~III.}
	In the absence of \creature, \hyperbard and \graph try to convey their message that representations matter to \colleague.
	\professor and \creature return, and \professor orders \colleague to eliminate \hyperbard.
	\emph{Act~V, Scene~I.}
	Having cremated \hyperbard, \colleague pours their ashes onto the graph dataset prepared earlier. 
	\graph mourns the death of their sovereign and sketches its implications.
	\emph{Act~V, Scene~II.}
	\creature wrestles with their experience in the Community.
	Instead of leaving in silence, they decide to tell their own story.

\subsection{The Dataset}
	The \ourdata dataset comprises $666$ graphs and hypergraphs:
	$18$ relational representations for each of $37$ plays by William Shakespeare (Fig.~\hyperlink{fig:rawdata}{1}). 
	From the TEI Simple XMLs provided by \folgertexts \cite{folger2022}, 
	for each play, we derive $6$ hypergraphs, $6$ clique expansions (i.e., interaction graphs), and $6$ star expansions (i.e., bipartite graphs) 
	that differ along $3$ dimensions (Tab.~\ref{tab:representations}, Fig.~\ref{fig:all}):
	\emph{semantic mapping}, \emph{granularity}, and \emph{expressivity}.
	As we show for \emph{Romeo and Juliet}, the representations we provide emphasize different aspects of the underlying raw data (Fig.~\ref{fig:representations:ce-scene}--\ref{fig:representations:se}, \ref{fig:representations:hg}), and they yield widely varying results even for simple measurements of character importance (Fig.~\hyperlink{fig:ranking:correlations}{7}--\ref{fig:rankings:dynamic}). 
	Thus, \ourdata \emph{enables} and \emph{demonstrates the need for} research on how representation choices impact the outputs and performance of graph learning, graph mining, and network analysis methods. 
	As such, it can be seen as an adaptation of \emph{multiverse analysis}, 
	originally developed in psychology~\cite{steegen2016increasing}, 
	to inherently relational data: 
	With \ourdata, we are introducing the idea of a \emph{graph data multiverse}.

\subsection{The Critique}
	\emph{The Community} is designed as a microcosm of \emph{our community}, 
	including all levels of academic seniority as well as common supporting roles.
	The characters \emph{inside} the Community exhibit cognitive, behavioral, and interaction patterns that frequently afflict people with corresponding roles in our community.
	The characters \emph{outside} the Community appear as their antidotes, challenging the status quo and engaging in free-spirited scientific inquiry.
	As the play progresses, \creature gets caught up between both worlds, 
	and we witness the force of community dynamics acting upon individuals that do not fit in.
	Examples of community phenomena featured in the play (there are many more):
	a struggling PhD student (\creature), 
	abuse of power and difficulties of criticism in hierarchical organizations (\professor), 
	administrative overload at the top of the pyramid (\professor and \seniorresearcher), 
	cynical resignation, disillusionment, and complicitness (\colleague), publish or perish (\deadlines), 
	academia vs. ``freedom'' (Community vs. forest), 
	mental health (\creature attempts \emph{suicide}), 
	uncomfortable viewpoints being shut down (\hyperbard is \emph{cremated}).

\clearpage

\section{Play Documentation}

\subsection{Inspirations}
\label{apx:play:inspirations}

The play deliberately adopts and adapts ideas and text fragments from Shakespeare's works and other popular texts. These are:
\begin{itemize}[label=--]
	\item Dramatis Person\ae:
	Three deadlines $\sim$ three witches from Shakespeare's \emph{Macbeth}
	\item Induction: Framing device used in Shakespeare's \emph{The Taming of the Shrew}
	\item Act~I, Scene~II, \checkednumber{\hyperlink{luther}{l.~32}}: 
	A phrase famously \emph{attributed} to Martin Luther
	\item Act II, Scene I, \checkednumber{\hyperlink{inquisition}{l.~127}}:
	Allusion to a series of sketches from Monty Python's Flying Circus
	\item Act II, Scene III, \checkednumber{\hyperlink{alltheworld}{ll.~159--179}}:
	Jon's speech from Shakespeare's \emph{As You Like It}
	\item Act II, Scene IV, \checkednumber{\hyperlink{faust}{ll.~184--191}}:
	Faust's speech from Goethe's Faust I
	\item Act III, Scene I, \checkednumber{\hyperlink{ariel}{ll.~303--316}}:
	Ariel's Song from Shakespeare's \emph{The Tempest}
	\item Act III, Scene I, \checkednumber{\hyperlink{hamlet}{ll.~319--332}}:
	Hamlet's monologue from Shakespeare's \emph{Hamlet}
	\item Act IV, Scene III, \checkednumber{\hyperlink{juliet}{l.~370}}:
	Juliet addressing Romeo in Shakespeare's \emph{Romeo and Juliet}
	\item Act IV, Scene III, \checkednumber{\hyperlink{fool}{ll.~401--402}}:
	Pieces from Jon's interactions in Shakespeare's \emph{As You Like It}
	\item Act V, Scene I, \checkednumber{\hyperlink{sonnet}{ll.~416--429}}:
	Shakespeare's \emph{Full Many a Glorious Morning Have I Seen} (Sonnet 33)
	\item Act V, Scene II, \checkednumber{\hyperlink{macbeth}{ll.~424--432}}:
	Macbeth's monologue from Shakespeare's \emph{Macbeth}
\end{itemize}

\subsection{Style}
\label{apx:play:style}

Our layout follows the Oxford Shakespeare from 1916 \cite{shakespeare1916complete} (whose text sometimes differs from the Folger Shakespeare underlying our data \cite{folger2022}, especially in the stage directions).
We adopt the basic language patterns characteristic of Shakespeare's plays, using primarily blank verse, i.e., non-rhyming verse in iambic pentameter with feminine endings allowed, 
but also prose and rhyming verse.
Our main character switches between blank verse and prose depending on their internal state.
Longer passages of rhyming verse occur in song and sonnet adaptations (see Section~\ref{apx:play:inspirations}); 
shorter passages of rhyming verse are scattered throughout the play. 
We generally use Modern American English, sprinkled with brief interludes of Old British English.

\end{document}